\documentclass[journal]{IEEEtran}

\usepackage{cite}

\ifCLASSINFOpdf
  \usepackage[pdftex]{graphicx}
\else
\fi

\usepackage{epstopdf}
\usepackage{amsmath}

\usepackage[caption=false,font=footnotesize]{subfig}

\usepackage{url}

\usepackage{booktabs}
\usepackage{multirow}
\usepackage{tablefootnote}
\usepackage[symbol]{footmisc}

\usepackage{amssymb}
\usepackage{color}

\newcommand{\vct}[1]{\boldsymbol{#1}}
\newcommand{\mat}[1]{\boldsymbol{#1}}
\newcommand{\tensor}[1]{\mathcal{#1}}

\newcommand{\para}[1]{\subsubsection{#1}}
\newcommand{\data}[1]{\textsf{#1}}
\newcommand{\venue}[1]{}

\def\ntloss{noise tolerance loss}
\def\Ntloss{Noise tolerance loss}
\def\NTLOSS{Noise Tolerance Loss}
\def\multi{multi-path}
\def\Multi{Multi-path}
\def\MULTI{Multi-Path}

\def\ourmethod{ROADMAP}

\newcommand{\onedot}{.}

\def\eg{\emph{e.g}\onedot}

\def\ie{\emph{i.e}\onedot}

\def\etc{\emph{etc}\onedot}

\def\etal{\emph{et al}\onedot}

\begin{document}
\title{Robust Unsupervised Video Anomaly Detection\\by {\MULTI} Frame Prediction}

\author{
    Xuanzhao~Wang,
    Zhengping~Che,~\IEEEmembership{Member,~IEEE,}
    Bo~Jiang,
    Ning~Xiao,
    Ke~Yang,\\
    Jian~Tang,~\IEEEmembership{Fellow,~IEEE,}
    Jieping~Ye,~\IEEEmembership{Fellow,~IEEE,}
    Jingyu~Wang,~\IEEEmembership{Member,~IEEE,}
    and~Qi~Qi
    \thanks{
        X.~Wang, N.~Xiao, and K.~Yang are with Beijing University of Post and Telecommunication, Beijing, China,
        and this work was done during their internship at Didi Chuxing, China.
        Z.~Che and B.~Jiang are with Didi Chuxing, Beijing, China.
        J.~Tang is with Midea Group, Beijing, China.
        J.~Ye is with Beike AI Tech, Beijing, China.
        J.~Wang and Q.~Qi are with Beijing University of Post and Telecommunication, Beijing, China.
    }
    \thanks{
        This work was supported in part by the National Key R\&D Program of China 2020YFB1807805, National Natural Science Foundation of China under Grants (62071067, 62001054, and 61771068), and Ministry of Education and China Mobile Joint Fund (MCM20200202 and MCM20180101).
    }
    \thanks{
        The corresponding authors are J.~Wang~(wangjingyu@bupt.edu.cn) and Z.~Che~(chezhengping@didiglobal.com).
    }
}

\markboth{Robust Unsupervised Video Anomaly Detection by {\MULTI} Frame Prediction}{IEEE Transactions on Neural Networks and Learning Systems}%

\IEEEspecialpapernotice{(Special Issue: Deep Learning for Anomaly Detection)}

\maketitle

\begin{abstract}
    Video anomaly detection is commonly used in many applications such as security surveillance and is very challenging.
A majority of recent video anomaly detection approaches utilize deep reconstruction models, but their performance is often suboptimal because of insufficient reconstruction error differences between normal and abnormal video frames in practice.
Meanwhile, frame prediction-based anomaly detection methods have shown promising performance.
In this paper, we propose a novel and robust unsupervised video anomaly detection method by frame prediction with proper design which is more in line with the characteristics of surveillance videos.
The proposed method is equipped with a {\multi} ConvGRU-based frame prediction network that can better handle semantically informative objects and areas of different scales and capture spatial-temporal dependencies in normal videos.
A {\ntloss} is introduced during training to mitigate the interference caused by background noise.
Extensive experiments have been conducted on the CUHK Avenue, ShanghaiTech Campus, and UCSD Pedestrian datasets, and the results show that our proposed method outperforms existing state-of-the-art approaches.
Remarkably, our proposed method obtains the frame-level AUROC score of 88.3\% on the CUHK Avenue dataset.

 \end{abstract}


\ifCLASSOPTIONpeerreview
\begin{center} \bfseries EDICS Category: 2-VSSA and 9-DLMA	\end{center}
\fi
\IEEEpeerreviewmaketitle

\section{Introduction}

\IEEEPARstart{V}{ideo} anomaly detection is an essential task in computer vision and used in many critical applications such as video surveillance~\cite{sodemann2012review,Sultani_2018_CVPR}, scene understanding~\cite{rhodes2006seecoast}, activity recognition~\cite{zhu2012context}, and road traffic analysis~\cite{li2016traffic}.
Given a video clip, the frame-level video anomaly detection aims at identifying the frames in which there exist events or behaviours different from expectations or regulations.
While an increasing amount of attention to developing effective video anomaly detection methods has been drawn from both academia and industry especially in recent years~\cite{Saligrama:2012:VAD:2354409.2355081,Lu:2013:AED:2586117.2586998,DBLP:conf/bmvc/XuRYSS15,Hasan_2016_CVPR,1609.08938,luo2017revisit,liu2018ano_pred,Ionescu_2019_CVPR},
detecting anomalous events in videos is still an extremely challenging task because of two major issues.
One is that the definition of an anomalous event is usually ambiguous, and its pattern varies a lot, highly depending on the applications and contexts.
For example, a car driving down the road is normal, but a car dashing into a park full of pedestrians is an anomaly;
a cloud of smoke in a building often indicates an abnormal fire, while it is normal in some hot spring areas.
Therefore, methods that are robust to different situations and environments for anomaly detection tasks are in great demand.
The other issue comes from the data imbalance between normal and abnormal samples in practice.
Anomalous events are rare and unpredictable in real-world scenarios.
Collecting and labelling abnormal videos is difficult and costly in terms of both time and manpower.
Thus, in most applications, detection methods are supposed to learn from normal data only and distinguish unseen and unbounded abnormal events from normal ones, i.e., in an unsupervised learning way.

Traditional video anomaly detection methods~\cite{kim2009observe,wu2010chaotic,Mahadevan.anomaly.2010,zhao2011online,YangCong:2011:SRC:2191740.2191781,Lu:2013:AED:2586117.2586998} carry the limited detection performance due to the poor discriminative power of the simple \textit{hand-crafted} features from prior domain knowledge.
Most state-of-the-art methods have achieved encouraging progress on video anomaly detection by leveraging the superiority of \textit{deep learning}, which can be grouped into three categories, if not a combination of them.
\textit{Classification-based} methods directly classify each video frame to be normal or not, which usually rely on massive annotations through supervised training approaches.
Among them, some methods~\cite{Sultani_2018_CVPR,Zhong_2019_CVPR} even require additional data and labels and others~\cite{Ionescu_2019_CVPR} merely focus on specific types of anomaly events.
As a result, the heavy consumption of labeling and confined scope of tasks have impeded the feasibility and generality of these methods for practical deployment.
Comparatively, as a more general way to tackle video anomaly detection, \textit{reconstruction-based} methods~\cite{Hasan_2016_CVPR,dimokranitou2017adversarial} attempt to mitigate the data labeling problem by reconstructing the video frames with small and large reconstruction errors for normal and abnormal frames, respectively. However, existing methods usually fail to learn the discrimination of reconstruction errors between normal and abnormal frames.
Recently, \textit{prediction-based} methods~\cite{liu2018ano_pred,ye2019anopcn} are introduced to determine the anomalies based on the frame prediction error. However, existing methods of this type deliver suboptimal results due to the insufficient modeling of temporal information and suffer from the inefficient training for employing adversarial techniques or additional losses.

Towards alleviating the above issue, in this paper, we propose \textbf{\ourmethod}, a novel and \textbf{r}obust unsupervised vide\textbf{o} \textbf{a}nomaly \textbf{d}etection method based on \textbf{m}ulti-path fr\textbf{a}me \textbf{p}rediction.
On one hand, noticing that the background in surveillance videos often keeps unchanged and semantically meaningful objects usually have varied sizes in the frames,
we design a {\multi} structure to efficiently skim the background and effectively extract features of different scales.
Our proposed method explicitly captures the temporal information in object and semantic motions via multiple convolutional gated recurrent
units~\cite{DBLP:journals/corr/BallasYPC15} of different resolutions with non-local blocks~\cite{NonLocal2018}.
On the other hand, we notice that consecutive video frames are full of randomly fluctuated noise, which heavily disturbs the prediction model and prevents the model from paying attention to the informative contents for differentiating anomalies.
Therefore, we introduce a {\ntloss} to effectively mitigate the impact of noise and boost the anomaly detection performance further.
The improvements on the proposed network structure and loss function are not mutually exclusive,
and intuitively, the noise tolerance loss and the multi-path ConvGRUs cooperate well: the noise tolerance loss reduce the interference of the noise pixels, and the meaningful changes between frames can be better extracted by the ConvGRU modules.
Our model is easy to train and robust to apply, since it does not need to use complicated variational methods or generative adversarial networks nor resort to additional losses such as optical flow loss or adversarial loss.
We evaluate our proposed {\ourmethod} on three benchmark datasets: CUHK Avenue, ShanghaiTech Campus, and UCSD Pedestrian.
The experimental results demonstrate that our model outperforms several strong baseline methods and achieves state-of-the-art performance.

The main contributions of this paper are as follows:
\begin{itemize}
    \item We propose a novel unsupervised video anomaly detection framework, namely {\ourmethod}, with suitable designs for robust performance in different scenarios.
    \item We equip {\ourmethod} with {\multi} ConvGRUs, which can handle informative parts of different scales, capture temporal relationship between frames, and pay less attention to static and background parts of the frame.
    \item We introduce a {\ntloss} to mitigate the interference caused by intrinsic noisy pixels in video frames, and the loss greatly improves the robustness and performance of the prediction-based anomaly detection.
    \item We conduct experiments on three publicly available video anomaly detection datasets. The superior performance of the proposed {\ourmethod} against many state-of-the-art baselines validates our design and shows its effectiveness.
\end{itemize}

\section{Related Work}
\subsection{Traditional Video Anomaly Detection}

For a long time in the past, anomaly detection methods are mainly based on hand-crafted features, such as histograms of oriented gradient (HOG) and histograms of oriented optical flow (HOF).
For example, Mehran \etal~\cite{mehran2009abnormal} constructs a social force model, and Li \etal~\cite{li2013anomaly} fits a mixture model for dynamic texture.
Hidden Markov models (HMMs)~\cite{hospedales2009markov,kratz2009anomaly}, Markov random fields (MRFs)~\cite{kim2009observe}, and Gaussian process models~\cite{li2015anomaly,cheng2015video} have also been used for modeling normal patterns and identifying anomalies.
Sparse reconstruction~\cite{YangCong:2011:SRC:2191740.2191781,zhao2011online,Lu:2013:AED:2586117.2586998,luo2019video} and locality sensitive hashing~\cite{zhang2016video} are also popular methods based on hand-crafted or shallow features.
However, these methods offer inferior performance because of the limited representational power of simple features.

\subsection{Deep Learning-Based Video Anomaly Detection}

With the rapid development of deep learning, pioneering works have raised impressive solutions to address video anomaly detection.
\textit{Classification-based} schemes decide the abnormal video frames by classifiers, e.g., Sabokrou \etal~\cite{sabokrou2018adversarially} applies end-to-end generative architecture for one-class classification for image and video anomaly detection, and Morais \etal~\cite{Morais_2019_CVPR} uses skeleton trajectories to learn pedestrian-related behavior. Besides, Ionescu {\etal} conducts a multi-stage learning strategy including clustering and one-versus-rest classification~\cite{ionescu2019detecting}, and later improves the performance by detecting the object of interests~\cite{Ionescu_2019_CVPR}. Furthermore, some works~\cite{Sultani_2018_CVPR,Zhong_2019_CVPR,ramachandra2020learning,Markovitz_2020_CVPR} require additional data or annotations to enhance the performance of classifiers. Nevertheless, the above approaches lack generality due to the costly data labeling or specific assumptions.
As for \textit{reconstruction-based} approaches, auto-encoders~\cite{Hasan_2016_CVPR,Zhao:2017:SAV:3123266.3123451,Gong_2019_ICCV,DBLP:conf/bmvc/XuRYSS15,Hasan_2016_CVPR,Zhao:2017:SAV:3123266.3123451,chong2017abnormal,luo2017revisit,luo2017remember,Gong_2019_ICCV,wu2019deep,wang2018detecting} and generative adversarial networks~\cite{dimokranitou2017adversarial,ravanbakhsh2017abnormal,ganokratanaa2020unsupervised} are widely adopted for video anomaly detection.
The method based on frame prediction is first proposed by Liu \etal~\cite{liu2018ano_pred} and has been extended by Ye \etal~\cite{ye2019anopcn}. They take consecutive video frames to predict the next frame and determine whether the next frame is abnormal by the prediction error. However, the U-Net architecture in~\cite{liu2018ano_pred} cannot thoroughly learn temporal information and using adversarial learning and additional optical flow loss makes the training inefficient.

\subsection{Video Frame Prediction}
Recently, video frame prediction has attracted more attention due to its potential applications in unsupervised video representation learning.
Shi \etal~\cite{xingjian2015convolutional} proposes a convolutional long short-term memory model for precipitation forecasting.
Mathieu \etal~\cite{DBLP:journals/corr/MathieuCL15} designs a multi-scale network with adversarial training to generate more natural future frames in videos.
Taking into account the characteristics of videos, Wang \etal~\cite{wang2018eidetic} designs a new long short-term memory structure (E3D-LSTM) and achieves excellent performance.
Moreover, several methods~\cite{chen2017video,van2017transformation} learn transformation-based frame generations instead of direct frame predictions.

\section{Methodology}
\label{sec:method}

\begin{figure*}[htb]
    \centering
    \includegraphics[width=.99\textwidth]{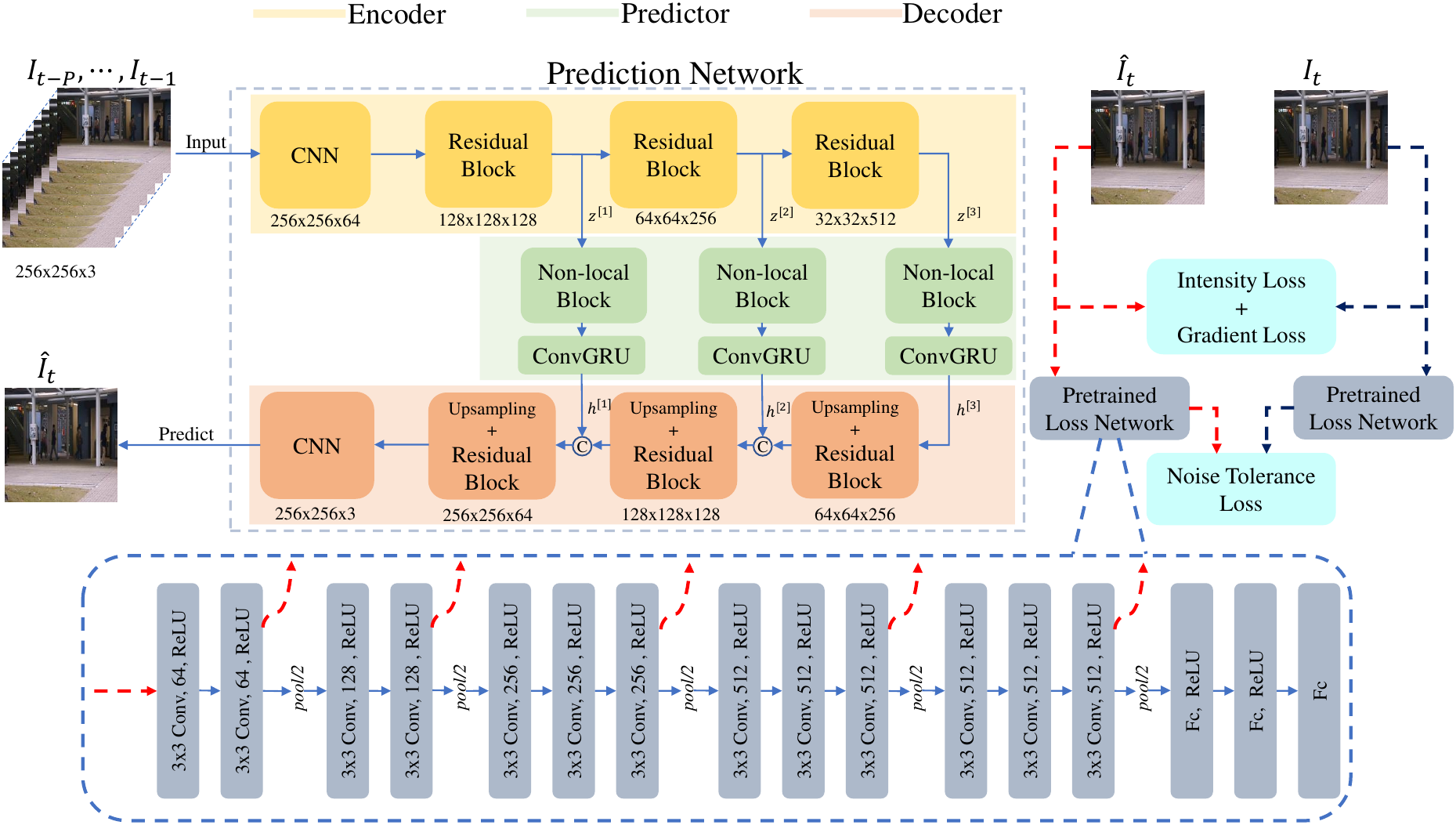}
    \caption{The overall architecture of the proposed model \textbf{\ourmethod}: \textbf{R}obust unsupervised vide\textbf{O} \textbf{A}nomaly \textbf{D}etection method based on \textbf{M}ulti-path fr\textbf{A}me \textbf{P}rediction.
    In order to determine whether the video frame $\mat{I}_t$ is abnormal, {\ourmethod} takes previous frames $\left(\mat{I}_{t-P}, \cdots, \mat{I}_{t-1}\right)$ as the input and predicts the next frame $\hat{\mat{I}_t}$. An anomaly score is calculated based on the difference between $\mat{I}_t$ and $\hat{\mat{I}_t}$.
    }
    \label{fig:architecture}
\end{figure*}

\subsection{Overview}
\label{sec:method-overall}

The frame-level video anomaly problem can be formally described as follows: In a video clip $\tensor{V}$ with $T$ consecutive frames $\mat{I}_1, \mat{I}_2, \cdots, \mat{I}_T$, each frame $\mat{I}_t$ is assigned with a binary label $y_t$ which indicates whether the content in the $t$-th frame is anomalous ($y_t = 1$) or not ($y_t = 0$).

Built upon the idea of anomaly detection by frame prediction~\cite{liu2018ano_pred}, our proposed method {\ourmethod} consists of a \textit{prediction network} $f(\cdot)$ and an \textit{assessment model} $g(\cdot)$ to predict frames based on the history and to determine anomalies based on the predictions, respectively.
Given $P$ consecutive frames $\mat{I}_{t-P}, \cdots, \mat{I}_{t-1}$, the prediction network generates the next frame $\hat{\mat{I}_t} = f \left( \mat{I}_{t-P}, \cdots, \mat{I}_{t-1} \right)$.
We apply a {\multi} encoder-decoder architecture with recurrent neural networks in order to capture both spatial and temporal dependencies of different scales for accurate prediction, and we introduce a {\ntloss} during network training to eliminate the influence of undesired noise.
The predicted frame embodies our expectations for the events about to happen.
Therefore, the difference between a predicted frame $\hat{\mat{I}_t}$ and its ground truth $\mat{I}_t$ should be small if $\mat{I}_t$ is normal.
If the difference between two frames is large, $\mat{I}_t$ is likely to contain anomaly events.
The assessment model compares the predicted frame with the ground truth and produces an anomaly score $S_t = g(\mat{I}_t, \hat{\mat{I}_t} ) \in [0, 1]$ by evaluating and normalizing the prediction quality.
This anomaly detection pipeline is conducted in a sliding window approach for all frames in the video.

In the remainder of this section, we first present our prediction network architecture of {\ourmethod} in Section~\ref{sec:method-multiscale}, introduce the {\ntloss} and overall objective function for training in Sections~\ref{sec:method-perceptual}~and~\ref{sec:method-objective}, and finally describe the anomaly assessment model in Section~\ref{sec:method-psnr}.

\begin{figure*}[tbh]
    \centering
    \includegraphics[width=.99\textwidth]{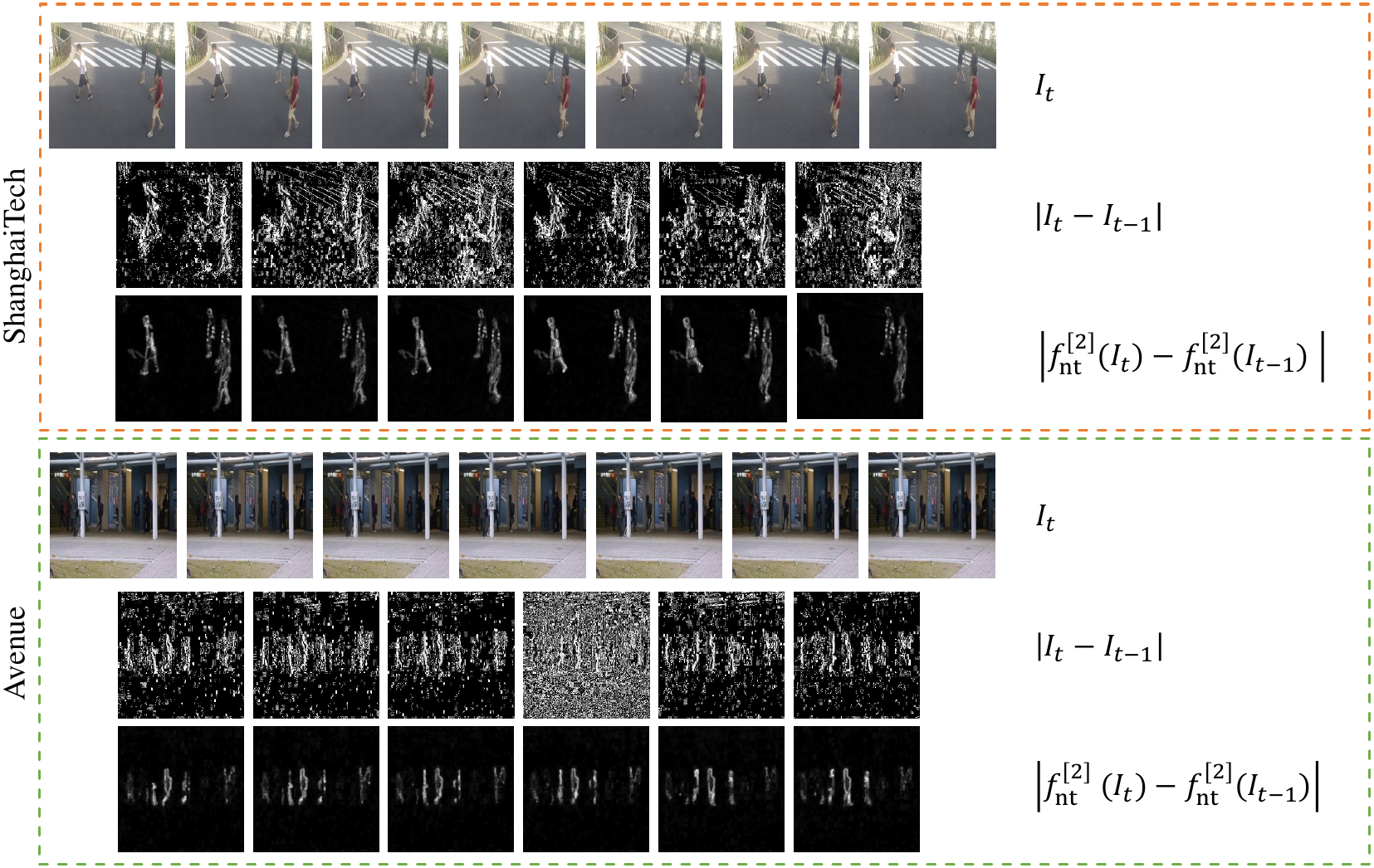}
    \caption{
        Examples of gradient noise on \data{ShanghaiTech} (top) and \data{Avenue} (bottom).
        Top rows: raw consecutive frames;
        middle rows: gradient of raw frames with much noise;
        bottom rows: gradient of the outputs from VGG16's second ReLU layer with less noise.
    }
    \label{fig:perceptual-example}
\end{figure*}

\subsection{{\MULTI} Frame Prediction Network}
\label{sec:method-multiscale}

The quality of the frame prediction heavily contributes to the performance of the anomaly detection.
To better fit the normal scenes in surveillance videos, we properly design the prediction network architecture with three parts: the \textit{encoder} $f_{\mathrm{enc}}\left(\cdot\right)$, the \textit{predictor} $f_{\mathrm{pre}}\left(\cdot\right)$, and the \textit{decoder} $f_{\mathrm{dec}}\left(\cdot\right)$. The overall architecture of the prediction network is shown in Fig.~\ref{fig:architecture}.

\para{Encoder}
The encoder is composed of several basic residual blocks with 2D convolutions and extracts multi-scale spatial features from the input frame.
These features at different scales are to be fed into different predictor paths afterwards.
Unlike standard residual blocks~\cite{he2016deep}, our residual blocks remove all batch normalization layers.
This modification keeps the flow of the extracted spatial information through the encoder and has been shown useful~\cite{lim2017enhanced}.
We apply the same encoder to all input frames.

\para{Predictor}
The predictor is equipped with $L=3$ parallel prediction modules (paths), each of which separately predicts the information of the next frame in the form of features in different resolutions.
Each prediction module consists of a \textit{non-local} block~\cite{NonLocal2018} and convolutional gated recurrent units (\textit{ConvGRUs}~\cite{DBLP:journals/corr/BallasYPC15}).
The upstream non-local block captures long-range dependencies at all positions in the image.
The ConvGRU naturally learns temporal patterns with its small receptive field focused on a local neighborhood of the output from the non-local block.
These two parts coherently enable each prediction module to incorporate both local and global spatial-temporal information.
Furthermore, prediction modules at different scales in parallel paths together build a richer hierarchy of complementary information.

\para{Decoder}
The decoder fuses features at different scales from different predictor modules one after another by constructing the output frame using upsampling and channel concatenation operations.
We perform the upsampling operation by nearest neighbor interpolation followed by residual blocks to eliminate checkerboard artifacts in the output~\cite{odena2016deconvolution}. Similar to those in the encoder, no residual blocks in the decoder have batch normalization layers.
The output of the decoder has the same shape with each input frame to the encoder.
To summarize, we predict the next video frame at time $t$ based on its $P$ previous frames by the following three steps.
First, from each frame $\mat{I}_{t'}$ where $t' = t-P, \cdots, t-1$, the encoder produces $L$ different states, \ie,
\begin{align}
    \left( \vct{z}_{t'}^{[1]}, \vct{z}_{t'}^{[2]}, \cdots, \vct{z}_{t'}^{[L]} \right) = f_{\mathrm{enc}}\left( \mat{I}_{t'} \right).
\end{align}
Second, each module of the predictor makes its own prediction based on the corresponding outputs from the $P$ encoders, \ie,
\begin{align}
    \vct{h}_{t}^{[l]} = f_{\mathrm{pre}}^{[l]} \left(  \vct{z}_{t-P}^{[l]},  \vct{z}_{t-P+1}^{[l]}, \cdots,  \vct{z}_{t-1}^{[l]} \right)
\end{align}
where $l=1, 2, \cdots, L$.
Finally, the decoder obtains the prediction by
\begin{align}
    \hat{\mat{I}_t} = f_{\mathrm{dec}}\left( \vct{h}_{t}^{[1]}, \vct{h}_{t}^{[2]}, \cdots, \vct{h}_{t}^{[L]} \right).
\end{align}

\para{Design principle}
Our proposed network structure is carefully designed for video anomaly detection in surveillance based on its data and application characteristics which have not been thoroughly utilized by related network before.
The predictor is not only supposed to generate plausible frames in normal circumstances but also needs to enlarge the gap between an abnormal frame and the prediction of it.
First, a majority of pixels in a frame are the background which is usually consistent with minor changes between consecutive frames.
Recovering the background is relatively easy but less helpful for recognizing abnormal events.
The design of the {\multi} architecture allows the background information to be passed through shallow prediction modules and pays more attention in deeper modules to semantically meaningful parts such as object motion and intrusion, in a residual connection way.
Second, in surveillance videos, the same object of interests, \eg, a person or a vehicle, may appear in extremely varied sizes due to its distance to the camera.
The parallel predictors are sensitive to different scales of features and effectively model both large and small objects.
Besides, temporal smoothness exists in most normal events and behaviors and is a key to distinguishing abnormal frames. It can be efficiently captured via the RNN blocks without any additional components or complicated designs~\cite{liu2018ano_pred}.

\subsection{{\NTLOSS}}
\label{sec:method-perceptual}
The frame prediction model is forced to recover the value of every pixel.
Ideally, background and static objects in consecutive frames keep unchanged, and the model only predicts informative changes such as object motions.
However, in practice, we always find much internal or interfered noise in frames captured by cameras.
Such noise causes irregular and unpredictable pixel fluctuations between adjacent frames, which dominate the training loss and significantly disturb the model training.
To demonstrate this phenomenon, we show two examples on the ShanghaiTech Campus and CUHK Avenue datasets in Fig.~\ref{fig:perceptual-example}.
As shown in the top rows, the differences of background between adjacent raw frames are visually imperceptible.
However, when we check the \textit{gradient image} of raw images, which is calculated by $| \mat{I}_{t} - \mat{I}_{t-1} |$ between two adjacent images $\mat{I}_{t-1}$ and $\mat{I}_{t}$, we notice that there is plenty of noise all over the place.
Moreover, the degree of noise can vary over different frames. For example, the fourth gradient image of \data{Avenue} in Fig.~\ref{fig:perceptual-example} is more noisy than others.
The irregular noise conceals real and informative changes in videos and interferes with the predictions.

To mitigate the interference caused by noise and make the prediction model more robust, we introduce our {\ntloss}, which is the perceptual loss~\cite{Johnson2016Perceptual} obtained by a pre-trained loss network.
Our work is the first to introduce it for eliminating irrelevant noise in video anomaly detection.\footnote[2]{The term ``perceptual loss'' used in existing video anomaly detection paper~\cite{Morais_2019_CVPR} refers to a different and irrelevant loss function.}
Similar losses have helped several computer vision tasks such as metric learning~\cite{dosovitskiy2016generating}, motion transfer~\cite{chan2019everybody}, style transfer~\cite{huang2017arbitrary} and face image editing~\cite{jo2019sc}.
While they mainly focus on generating high-quality visual results by extracting and reconstructing high-level features,
our method uses the loss to eliminate the interference of the noise in captured videos so that the anomaly detection results are robust to the unpredictable noise.

In order to apply the noise tolerance loss, when training the prediction network, we simultaneously feed the predicted and ground truth frame into the loss network and then calculate the loss based on the hidden layer outputs.
We use the VGG16 network trained on ImageNet~\cite{deng2009imagenet} as the loss network, and its intermediate outputs carry much semantic information while filtering out irrelevant noise.

As shown in the bottom rows in Fig.~\ref{fig:perceptual-example}, on the ShanghaiTech Campus and CUHK Avenue datasets, the gradient noise of the output from the loss network (VGG16) is significantly eliminated compared with the original frame.
This implies that the information from these hidden layer outputs can supervise the model training and make the model pay more attention to the semantic changes while ignoring the noise.
Therefore, we use a weighted $\ell_1$ distance of the outputs from some intermediate layers of the loss network as the {\ntloss}.
Given $V = \{2, 4, 7, 10, 13\}$ as the set of selected layers of the loss network, and hyperparameters $\alpha_{v}$ controlling the strength on the loss of $v$-th layer, the {\ntloss} is defined by
\begin{equation}
    \mathcal{L}_{\mathrm{nt}}(\mat{I}_t, \hat{\mat{I}_t}) =
    \sum_{v \in V} { \alpha_{v} {\left\|f_{\mathrm{nt}}^{[v]}(\mat{I}_t)-f_{\mathrm{nt}}^{[v]}(\hat{\mat{I}_t})\right\|}_{1} }
    \label{eq:vgg}
\end{equation}
where $f_{\mathrm{nt}}^{[v]}(\cdot)$ denotes the output from the $v$-th layer of the loss network.

The noise tolerance loss fits well with the prediction-based method:
The random and internal noise in the captured videos is ubiquitous and by no means it can be predicted accurately.

\subsection{Training Frame Predictor}
\label{sec:method-objective}
As using the {\ntloss} alone may lose useful details in raw frames, we include the intensity loss $\mathcal{L}_{\mathrm{int}}$ and the gradient difference loss $\mathcal{L}_{\mathrm{gd}}$~\cite{DBLP:journals/corr/MathieuCL15} between the predicted frame $\hat{\mat{I}_t}$ and its ground-truth frame $\mat{I}_t$ for training as well.

Our intensity loss is the $\ell_2$ distance between two frames in the intensity space.
Given the two frames of height $H$ by width $W$, the intensity loss is defined by
\begin{equation}
    \mathcal{L}_{\mathrm{int}}(\mat{I}_t, \hat{\mat{I}_t}) =  \sum_{i, j} {\left\|\mat{I}_t(i, j) - \hat{\mat{I}_t}(i, j)\right\|}_2
\end{equation}
where $1\le i \le H$ and $1\le j \le W$. $\mat{I}_t(i, j)$ and $\hat{\mat{I}_t}(i, j)$ denote the RGB values of the $(i,j)$-th pixel in $\mat{I}_t$ and $\hat{\mat{I}_t}$, respectively.

The gradient difference loss is measured by the $\ell_1$ distance in both vertical and horizontal directions.
Specifically, the gradient differences are defined by
\begin{align}
    & \Delta_{\mathrm{gd}}^{[i]}(\mat{I}, i, j) = \left|\mat{I}(i,j)-\mat{I}(i-1,j)\right|, \\
    & \Delta_{\mathrm{gd}}^{[j]}(\mat{I}, i, j) = \left|\mat{I}(i,j)-\mat{I}(i,j-1)\right|,
\end{align}
and the gradient difference loss is calculated by
\begin{align}
     \nonumber \mathcal{L}_{\mathrm{gd}}(\mat{I}_t, \hat{\mat{I}_t}) =&
     \sum_{i, j} {\Big\| \Delta_{\mathrm{gd}}^{[i]}(\mat{I}_t, i, j) - \Delta_{\mathrm{gd}}^{[i]}(\hat{\mat{I}_t}, i, j) \Big\|}_1 \\
     & + \sum_{i, j} {\Big\| \Delta_{\mathrm{gd}}^{[j]}(\mat{I}_t, i, j) - \Delta_{\mathrm{gd}}^{[j]}(\hat{\mat{I}_t}, i, j) \Big\|}_1.
\end{align}

The overall objective function for network training is a weighted combination of all three losses with hyperparameters $\lambda_{\mathrm{int}}$, $\lambda_{\mathrm{gd}}$, and $\lambda_{\mathrm{nt}}$, defined by
\begin{align}
\nonumber    \mathcal{L}(\mat{I}_t, \hat{\mat{I}_t}) = &
    \lambda_{\mathrm{int}} \mathcal{L}_{\mathrm{int}}(\mat{I}_t, \hat{\mat{I}_t})
    + \lambda_{\mathrm{gd}} \mathcal{L}_{\mathrm{gd}}(\mat{I}_t, \hat{\mat{I}_t}) \\
    & + \lambda_{\mathrm{nt}} \mathcal{L}_{\mathrm{nt}}(\mat{I}_t, \hat{\mat{I}_t}).
    \label{eq:loss}
\end{align}

\subsection{Calculating Anomaly Score}
\label{sec:method-psnr}
To infer whether a frame is anomalous or not, our assessment model computes an anomaly score based on \textit{peak signal-to-noise ratio} (PSNR), similar to existing works~\cite{DBLP:journals/corr/MathieuCL15,liu2018ano_pred} for image quality assessment.
We first calculate the \textit{mean squared error} (MSE) between the ground-truth frame $\mat{I}_t$ and the predicted frame $\hat{\mat{I}_t}$ by
\begin{equation}
    \mathrm{MSE} ( \mat{I}_t, \hat{\mat{I}_t}  ) = \frac{1}{HW} \sum_{i, j} {\left\|\mat{I}_t(i, j) - \hat{\mat{I}_t}(i, j)\right\|}_2^2
\end{equation}
where $H$ and $W$ are the height and the width of frames, respectively, and $\mat{I}_t(i, j)$/$\hat{\mat{I}_t}(i, j)$ denotes the RGB values of the $(i,j)$-th pixel in $\mat{I}_t$/$\hat{\mat{I}_t}$.
The PSNR value $R_t$ between the two frames is calculated by
\begin{equation}
    R_t = 10 \log_{10} \frac{ \mathrm{MAX}_{\mat{I}_t}^2}{\mathrm{MSE} ( \mat{I}_t, \hat{\mat{I}_t} ) }
\end{equation}
where $\mathrm{MAX}_{\mat{I}_t}$ denotes the maximum possible pixel value of $\mat{I}_t$.
Finally, we calculate the anomaly score $S_t$ of the $t$-th frame $\mat{I}_t$ by normalizing the PSNR values of all $T$ frames in one video by
\begin{equation}
    S_t= 1 - \frac{ R_t - \min_{t'} R_{t'} }{\max_{t'} R_{t'} - \min_{t'} R_{t'}},
\end{equation}
where $t'=1, 2, \cdots, T$.
A higher score $S_t$ indicates that the frame $\mat{I}_t$ is more likely to be anomalous.

\section{Experiments}

\subsection{Datasets}
We evaluated the proposed {\ourmethod} on three publicly available datasets: the CUHK Avenue dataset~\cite{Lu:2013:AED:2586117.2586998}, the ShanghaiTech dataset~\cite{luo2017revisit}, and the UCSD Pedestrian dataset~\cite{Mahadevan.anomaly.2010}.

The CUHK Avenue dataset (\data{Avenue}) is one of the most widely used benchmarks for video anomaly detection. It contains 16 training videos and 21 testing videos with 47 abnormal events captured in CUHK campus avenue.
Each video is about 1 minute long with a resolution of 640$\times$360.
The anomalies include some unusual actions of people in the clips, \eg, running, loitering, and throwing.
There are 11457 normal and 3867 abnormal frames in the testing set.

The ShanghaiTech Campus dataset (\data{ShanghaiTech}) is among the largest datasets for video anomaly detection.
Unlike the other datasets, the video clips are from 13 different cameras with various lighting conditions and camera angles.
There are 330 training videos and 107 test videos with 130 abnormal events.
The resolution of each video frame is 856$\times$480.
There are 23465 normal and 17326 abnormal frames in the testing set.

The UCSD Pedestrian dataset has two different subsets (\data{Ped1} and \data{Ped2}).
\data{Ped1} includes 34 training videos and 36 testing videos with 40 abnormal events, and each video has 200 grayscale frames with a resolution of 238$\times$158.
\data{Ped2} contains 16 training videos and 12 testing videos with 12 abnormal events, and each video has about 170 grayscale frames with a resolution of 360$\times$240.
All abnormal cases are intrusions of bicycles and cars.
There are 3155 normal and 4045 abnormal frames in the testing set of \data{Ped1}, and there are 362 normal and 1648 abnormal frames in the testing set of \data{Ped2}.

\subsection{Evaluation Metrics}
Following previous works~\cite{1609.08938,liu2018ano_pred,Ionescu_2019_CVPR}, we computed the frame-level \textit{receiver operating characteristics} (ROC) and use the \textit{area under the ROC curve} (AUROC) score as the main evaluation metric.
We also calculated the \textit{area under the precision-recall curve} (AUPRC) score and the F1 score (the harmonic mean of precision and recall) as additional evaluation metrics.
A higher score (AUROC, AUPRC, and F1) indicates better anomaly detection performance.
We first obtained the anomaly scores for all video frames and then calculated the scores globally for each dataset.

\subsection{Implementation Details}
\label{sec:exp-imp}

All frames were resized to 256$\times$256, and the intensity of all pixels was normalized to $[-1, 1]$.
We used $P=8$ consecutive frames to predict the next frame.
For prediction network training, we followed Johnson \etal~\cite{Johnson2016Perceptual} and set $(\alpha_{2},\alpha_{4},\alpha_{7},\alpha_{10},\alpha_{13})=(0.1,1,10,10,10)$ in Equation~\ref{eq:vgg}.
We set the hyperparameters $\lambda_{\mathrm{int}} = 1$ and $\lambda_{\mathrm{gd}} = 1$ in Equation~\ref{eq:loss} for all datasets.
Based on the data characteristics of each dataset, we followed common practice~\cite{liu2018ano_pred,Morais_2019_CVPR} and set the hyperparameter $\lambda_{\mathrm{nt}}$ differently: $1$ for \data{Avenue}, $10$ for \data{ShanghaiTech}, and $0$ for \data{Ped1} and \data{Ped2}.
The batch size, learning rate, and weight decay were set to 4, 0.0003, and 0.0001, respectively.
The prediction model was trained for 50 epochs with \texttt{AdamW}-based stochastic gradient descent method~\cite{DBLP:conf/iclr/LoshchilovH19}.
To fully train deeper modules, we fixed the weights of the first and second prediction modules, \ie, $f_{\mathrm{pre}}^{[1]}(\cdot)$ and $f_{\mathrm{pre}}^{[2]}(\cdot)$, after the 30th and 40th epochs, respectively.
For the loss network, we used the pretrained VGG16 network from the \texttt{TorchVision}~\cite{torchvision} package.
Our model was implemented using \texttt{PyTorch} on a machine with an Intel E5-2630 CPU, 45GB RAM, and an NVIDIA Telsa P40 GPU.

\subsection{Results}
\label{sec:exp-results}
We compared {\ourmethod} with several state-of-the-art video anomaly detection baselines trained with only normal videos and focused on the standard unsupervised learning settings.
The evaluation results of frame-level anomaly detection on all datasets by the AUROC score are presented in Table~\ref{tab:scores}.

\begin{table}[h]
    \centering
    \caption{AUROC scores of the anomaly detection results.}
    \label{tab:scores}
    {
        \begin{tabular}{lcccc}
            \toprule
            Method & \data{Avenue} & \data{ShanghaiTech} & \data{Ped1} & \data{Ped2}\\
            \midrule
            MPPCA~\cite{kim2009observe}{\venue{CVPR2009}} & -- & -- & 59.0\% & 69.3\%\\
            MPPC+SFA~\cite{Mahadevan.anomaly.2010}{\venue{CVPR2010}} & -- & -- & 66.8\% & 61.3\%\\
            MDT~\cite{Mahadevan.anomaly.2010}{\venue{CVPR2010}} & -- & -- & 81.8\% & 82.9\%\\
            Conv-AE~\cite{Hasan_2016_CVPR}{\venue{CVPR2016}} & 80.0\% & 60.9\% & 75.0\% & 85.0\%\\
            Del \etal~\cite{1609.08938}{\venue{ECCV2016}} & 78.3\% & -- & -- & --\\
            Conv-LSTM-AE~\cite{luo2017remember}{\venue{ICME2017}} & 77.0\% & -- & 75.5\% & 81.1\%\\
            Unmasking~\cite{Ionescu_2017_ICCV}{\venue{ICCV2017}} & 80.6\% & -- & 68.4\% & 82.2\%\\
            Hinami \etal~\cite{DBLP:journals/corr/abs-1709-09121}{\venue{ICCV2017}} & -- & -- & -- & 92.2\%\\
            Stacked RNN~\cite{luo2017revisit}{\venue{ICCV2017}} & 81.7\% & 68.0\% & -- & 92.2\%\\
            Liu \etal~\cite{liu2018ano_pred}{\venue{CVPR2018}} & 85.1\% & 72.8\% & {83.1\%} & 95.4\%\\
            Wang \etal~\cite{wang2018detecting}{\venue{MM2018}} & 85.3\% & -- & 77.8\% & 96.4\% \\
            Davide \etal~\cite{Abati_2019_CVPR}{\venue{CVPR2019}} & -- & 72.8\% & -- & 95.4\%\\
            MPED-RNN~\cite{Morais_2019_CVPR}{\venue{CVPR2019}}\tablefootnote[3]{Results on a subset with only person-related abnormal events.} & 86.3\% & 75.4\% & -- & --\\
            Ionescu \etal~\cite{Ionescu_2019_CVPR}{\venue{CVPR2019}}\tablefootnote[9]{AUROC scores are recalculated by Feng \etal~\cite{object_centrci_VAD} based on the original results for fair comparison.} & 86.5\% & \textbf{78.5\%} & -- & --\\
            DeepOC~\cite{wu2019deep}{\venue{TNNLS2019}} & 86.6\% & -- & 75.5\% & \textbf{96.9\%} \\
            AnoPCN~\cite{ye2019anopcn} & 86.2\% & 73.6\% & -- & 96.8\% \\
            Nguyen \etal~\cite{Nguyen_2019_ICCV}{\venue{ICCV2019}} & {86.9}\% & -- & -- & 96.2\%\\
            MemAE~\cite{Gong_2019_ICCV}{\venue{ICCV2019}} & 83.3\% & 71.2\% & -- & 94.1\%\\
            sRNN-AE~\cite{luo2019video}{\venue{TPAMI2019}} & 83.5\% & 69.6\% & -- & 92.2\%\\
            Markovitz \etal~\cite{Markovitz_2020_CVPR}{\venue{CVPR2020}} & -- & 76.1\% & -- & -- \\
            \cmidrule{1-5}
            \textbf{{\ourmethod}} (ours) & \textbf{88.3\%} & \textbf{76.6\%} & \textbf{83.4}\% & \textbf{96.3\%}\\
            \bottomrule
        \end{tabular}
    }
\end{table}

\para{Avenue}
On the CUHK Avenue dataset, our proposed method {\ourmethod} outperformed all compared strong baselines with an AUROC score of 88.3\%, which is by 3.2 percentage points greater than that of the frame prediction-based anomaly detection baseline~\cite{liu2018ano_pred}.
This is as far as we know the best result in terms of the frame-level AUROC score on all testing videos in this dataset.
Notably, the results from Ionescu \etal~\cite{ionescu2019detecting,Ionescu_2019_CVPR} (88.9\% and 90.4\%) were calculated by a different metric in their papers,
and the recalculated frame-level AUROC score based on their best original anomaly detection results is 86.5\%~\cite{object_centrci_VAD}, which is by 1.8 percentage points smaller than that of our method with the same evaluation metric.

\para{ShanghaiTech}
Our method {\ourmethod} achieved a frame-level AUROC score of 76.6\%, which is by 3.8 percentage points greater than that of the frame prediction-based anomaly detection baseline~\cite{liu2018ano_pred} and second only to 78.5\% of Ionescu \etal~\cite{Ionescu_2019_CVPR}.
As Ionescu \etal~\cite{Ionescu_2019_CVPR} uses an object detection-based method for anomaly detection, its performance highly depends on the outputs from its object detection algorithm.
Therefore, the detection-based methods can not determine anomalous events related to unseen or unexpected objects, which is common in unbounded real-world anomaly detection applications and is rare in this testing dataset.
Similarly, Markovitz \etal~\cite{Markovitz_2020_CVPR} is aided by pretrained pose estimators and thus limited to detecting only human-related anomaly events.
In contrast, our frame prediction-based method does not have such limitations and is robust when applied to a variety of scenarios.
Compared with other recent strong baselines in Table~\ref{tab:scores}, {\ourmethod} gained advantages of more than 1 percentage point in terms of the AUROC score on the ShanghaiTech Campus dataset.

\para{UCSD Ped1 and Ped2}
These two datasets are relatively small, with a significantly low frame resolution of 158$\times$238.
As we found that the {\ntloss} was little helpful due to the grayscale and low-resolution properties, we took away it during training ($\lambda_{\mathrm{nt}} = 0$).
Even in its incomplete form,
our proposed method {\ourmethod} achieved frame-level AUROC scores of 83.4\% on \data{Ped1} and 96.3\% on \data{Ped2} which are among the best.
The results on these two datasets validated the effectiveness of the designed {\multi} architecture itself.

\subsection{Ablation Studies}

In this section, we investigate each component in the proposed {\ourmethod} and validate their effectiveness in detail.
The ablation studies were conducted on all four datasets, and the full results are shown in Table~\ref{tab:ablation}.

\begin{table}[h]
    \centering
    \caption{
        The AUROC scores of the ablation studies.
    }
    \label{tab:ablation}
    {
        \begin{tabular}{l@{\ }c@{\ }c@{\ \ }c@{\ \ }c@{\ \ }c}
            \toprule
            & \multicolumn{1}{c}{Baseline\tablefootnote[4]{Baseline refers to the method with only intensity and gradient losses in Liu \etal~\cite{liu2018ano_pred}, in which the U-Net can be seen as a simplified 2D {\multi} (multi-scale) network architecture.}} & \multicolumn{4}{c}{Ours}\\
            \cmidrule(lr){2-2} \cmidrule(lr){3-6}
            ConvGRU & -- & $\checkmark$ & $\checkmark$ & $\checkmark$& $\checkmark$\\
            {\Multi} Network & * & -- & $\checkmark$ & $\checkmark$& $\checkmark$\\
            Non-local Block & -- & -- & -- & $\checkmark$& $\checkmark$\\
            {\NTLOSS} & -- & -- & -- & -- & $\checkmark$\\
            \cmidrule{1-6}
            \data{Avenue} & 82.8\% & 82.4\% & 86.1\% & 86.7\%& \textbf{88.3\%}  \\
            \data{ShanghaiTech} & -- & 72.9\% & 74.1\% & 74.5\%& \textbf{76.6\%}  \\
            \data{Ped1} & -- & 81.7\% & 84.3\% & \textbf{84.9\%} & --  \\
            \data{Ped2} & -- & 94.4\% & 95.4\% & \textbf{96.4\%} & --  \\
            \bottomrule
        \end{tabular}
    }
\end{table}

\para{{\Multi} network and ConvGRU}
We took the ablated model from Liu \etal~\cite{liu2018ano_pred} trained with only the intensity and gradient losses as the baseline on \data{Avenue}.
While having no complicated optical flow or adversarial learning components, this baseline still obtained a competitive AUROC score of 82.8\%.
First, we inserted one ConvGRU prediction module at the bottom of the U-Net and removed the shortcut concatenations in U-Net to get a pure single-path network.
The modified network architecture could be seen as the one in Fig.~\ref{fig:architecture}, with only the last prediction module and no non-local block.
As the original shortcuts could have suppressed gradient vanishing and passed information directly, the modified model suffered a decline in its performance as expected.
Thanks to the added ConvGRU part which captured temporal information explicitly, the AUROC score dropped to 82.4\% with a loss of only 0.3 percentage points on \data{Avenue}.
However, after we added back {\multi} connections equipped with ConvGRUs, the AUROC score was significantly boosted to 86.1\% with an improvement of 3.7 percentage points.
On other datasets, the ablated baseline is not available, but the improvement by {\multi} structure is still clear, ranging from 1.0 percentage point (on \data{Ped2}) to 2.6 percentage points (on \data{Ped1}).
The results demonstrate that the prediction fusion of different scales is extremely effective, especially with recurrent neural networks that capture complemented temporal information.

\para{Non-local block}
We further added a non-local block before the ConvGRU in each prediction module to better capture the long-range spatial information.
This modification helped the model obtain an improvement of 0.6 percentage points in terms of the AUROC score, from 86.1\% to 86.7\% on \data{Avenue},
and the ablated model performed the second best among compared strong baselines in Table~\ref{tab:scores}, which indicates that non-local blocks are essential components which enhance the learning ability and discriminative power of our model.
On other datasets, the model with non-local block gained from 0.4 to 1.0 percentage point improvement.

\para{{\Ntloss}}
In Section~\ref{sec:method-perceptual}, we have illustrated that noise causes significant interference to prediction and may impede anomaly detection models.
Without changing the network architecture, we added the {\ntloss} for training.
The complete {\ourmethod} method achieved an AUROC score of 88.3\% on \data{Avenue} with a substantial gain of 1.6 percentage points, and an AUROC score of 76.6\% on \data{ShanghaiTech} with a substantial gain of 2.1 percentage points.
As we mentioned previously, the {\ntloss} was little helpful to the grayscale and low-resolution videos in the UCSD Pedestrian datasets and we omitted the results.
The results on the two large and challenging datasets validate our assumption that mitigating interference caused by the indiscernible noise is critical to robust video anomaly detection and show the effectiveness of using the {\ntloss}.

\begin{figure*}[t]
    \centering
    \hfill
    \subfloat[\data{Avenue} Testing Video 13]{
        \begin{minipage}[t]{0.485\textwidth}
        \centering
        \includegraphics[width=1.0\linewidth]{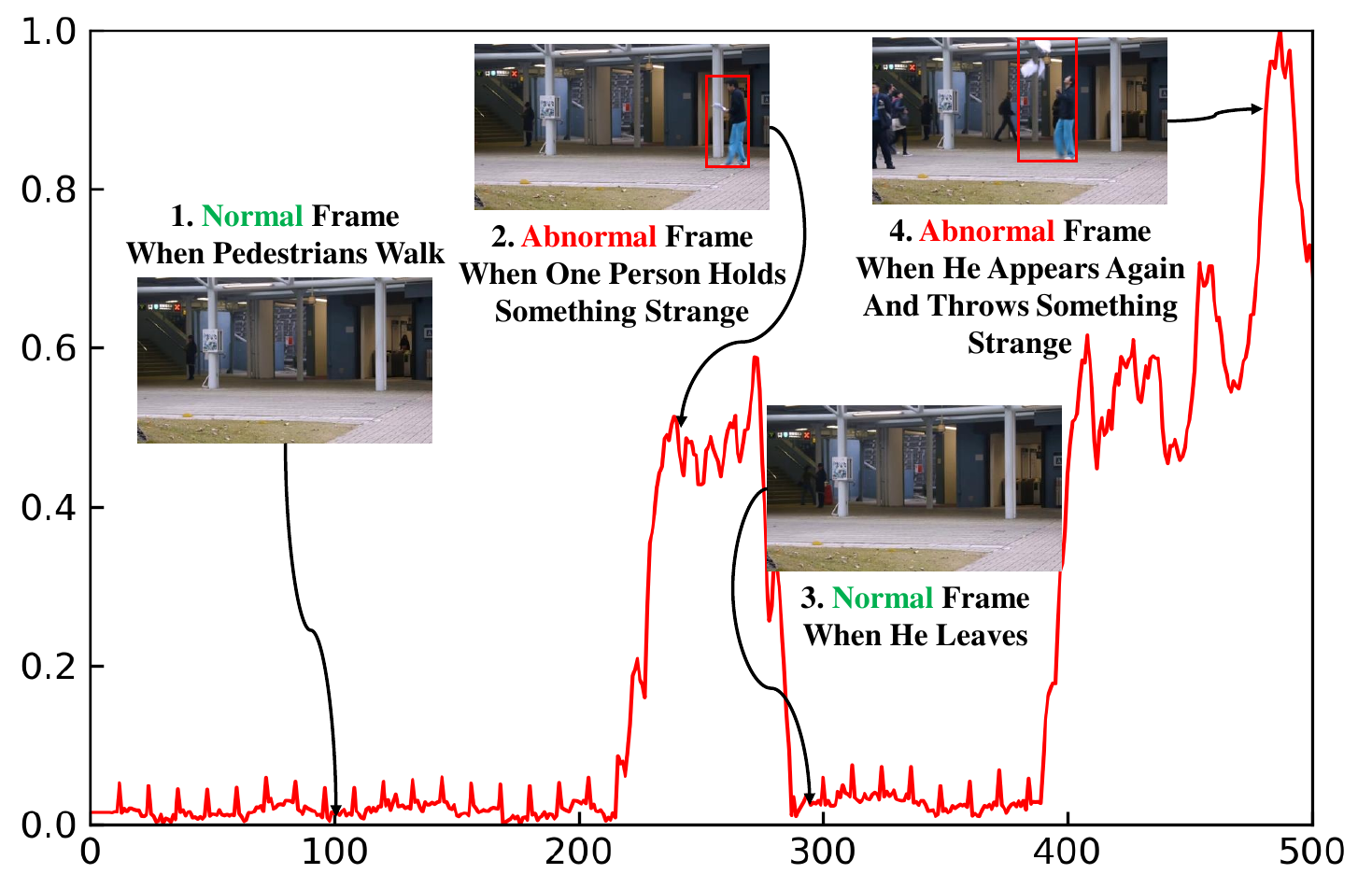}
        \end{minipage}
    }
    \hfill
    \subfloat[\data{Avenue} Testing Video 15]{
        \begin{minipage}[t]{0.485\textwidth}
        \centering
        \includegraphics[width=1.0\linewidth]{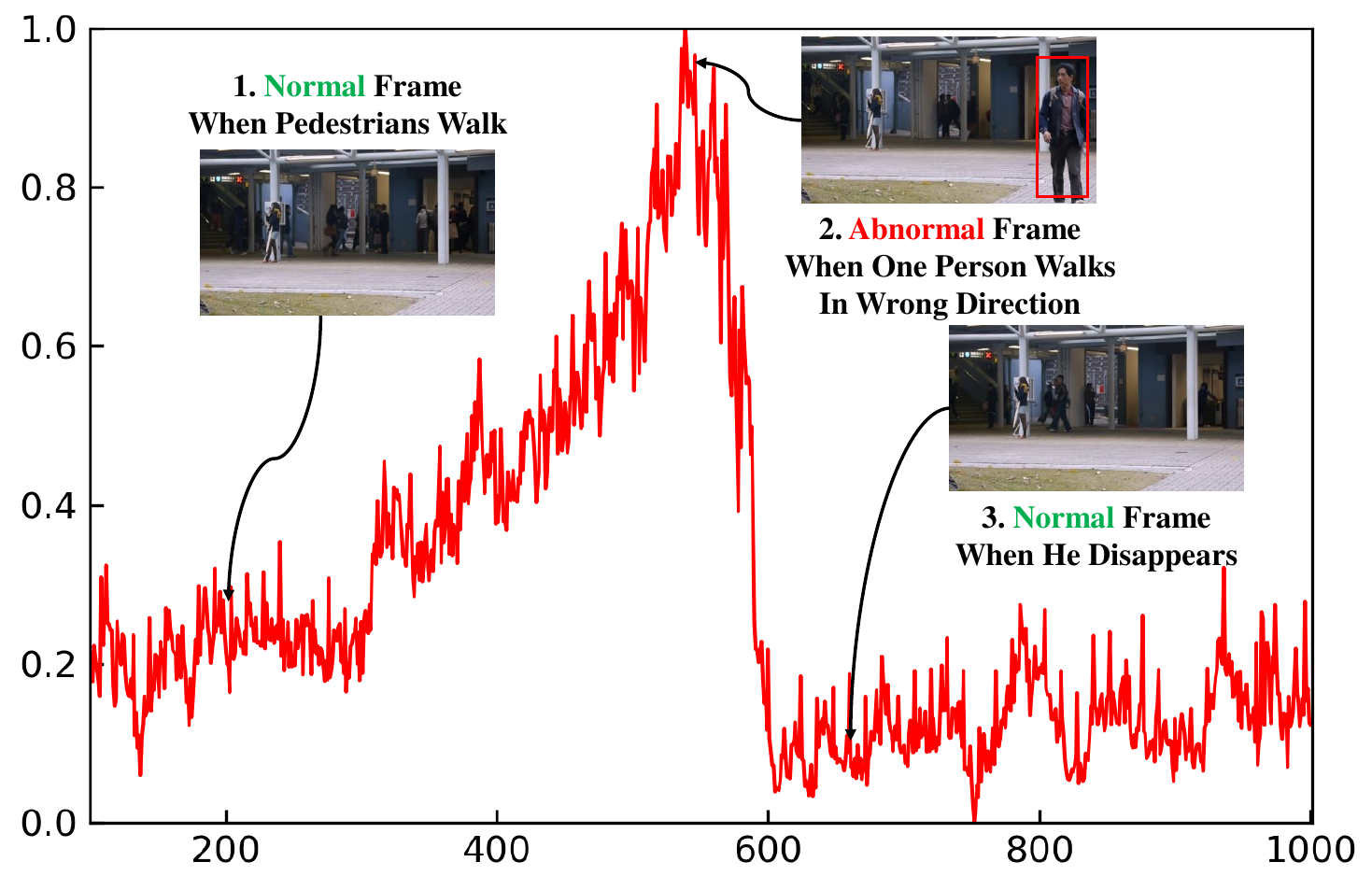}
        \end{minipage}
    }\hfill $ $ \\
    \hfill
    \subfloat[\data{ShanghaiTech} Testing Video 01-0134]{
        \begin{minipage}[t]{0.485\textwidth}
        \centering
        \includegraphics[width=1.0\linewidth]{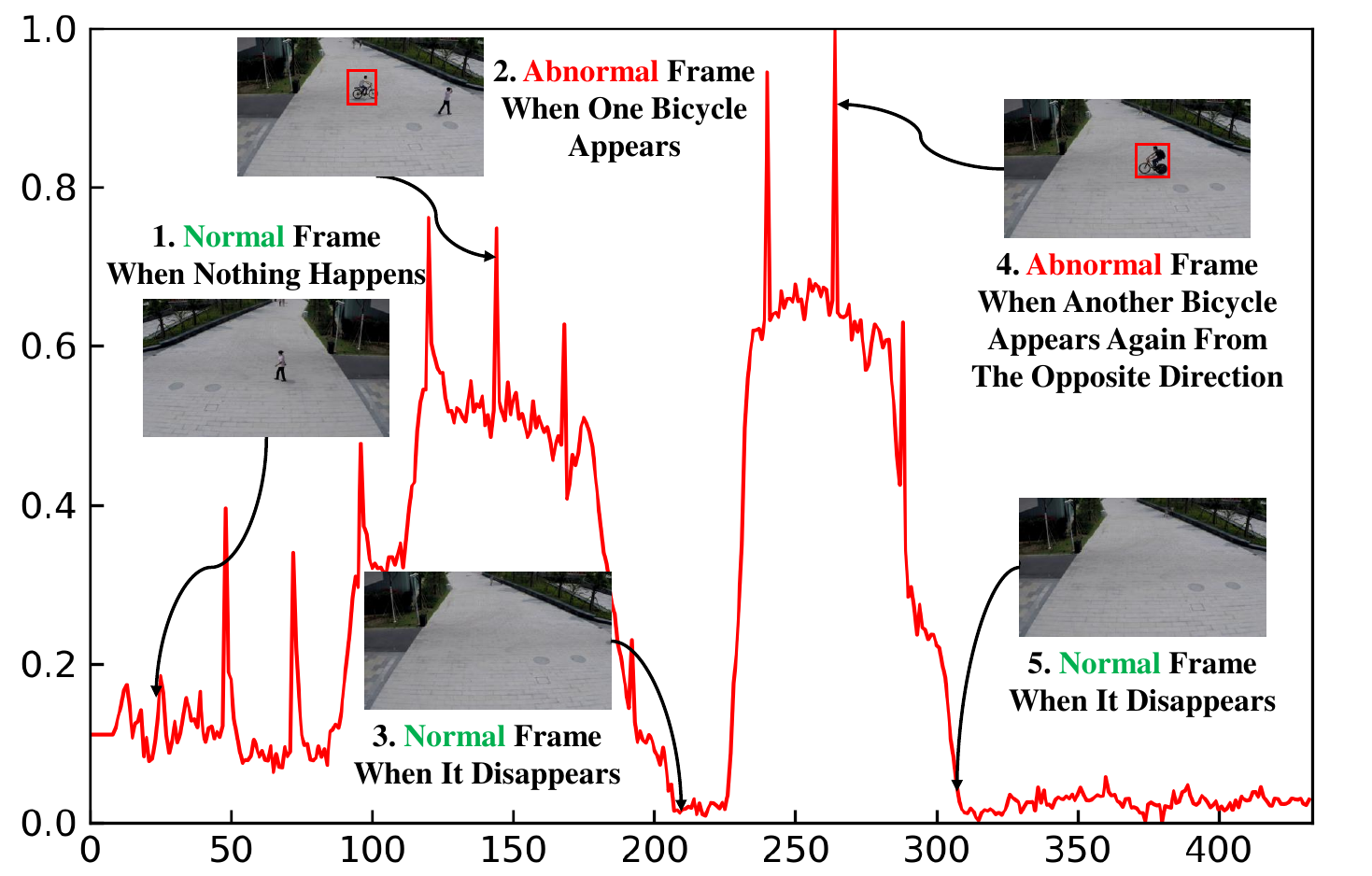}
        \end{minipage}
    }
    \hfill
    \subfloat[\data{Ped2} Testing Video 2]{
        \begin{minipage}[t]{0.485\textwidth}
        \centering
        \includegraphics[width=1.0\linewidth]{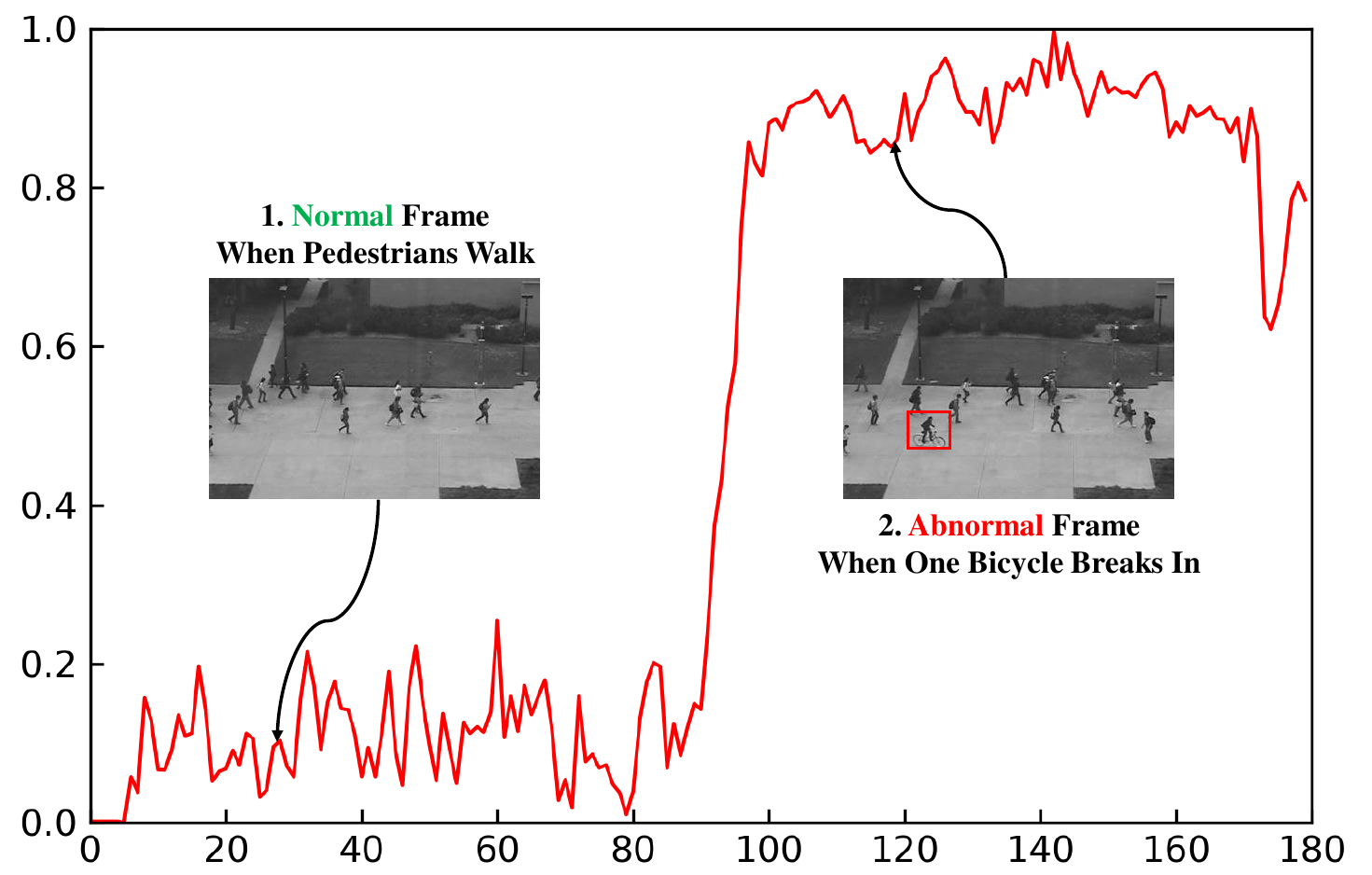}
        \end{minipage}
    } \hfill $ $
    \caption{Examples of anomaly score curves obtained by {\ourmethod} and representative video frames.}
    \label{fig:avenue-vis}
\end{figure*}

\subsection{Model and Result Investigations}

\para{Comparisons to related architectures}
While the results in Section~\ref{sec:exp-results} have shown the state-of-the-art performance of our approach,
we also calculated the gap between average scores of normal and abnormal frames to validate the superiority of its architecture.
The score gap $\Delta_{\mathrm{S}}$ on dataset $\tensor{D}$ is defined by
\begin{equation}
\Delta_{\mathrm{S}}
= \sum_{\tensor{V} \in \tensor{D}} \sum_{t \in \{t | \mat{I}_t \in \tensor{V} \} } (2 y_t - 1) S_t,
\end{equation}
and a higher value of $\Delta_{\mathrm{S}}$ indicates a more robust network for distinguishing normal and abnormal events.
Table~\ref{tab:gap} shows that our {\ourmethod} obtained larger score gaps consistently than existing simplified architectures (multi-scale, auto-encoder, U-Net, \etc) and validates the effectiveness of our design.

\begin{table}[h]
    \centering
    \caption{
        The gap $\Delta_{\mathrm{S}}$ between average anomaly scores of normal and abnormal frames.
    }
    \label{tab:gap}
    {
        \begin{tabular}{lcccc}
            \toprule
            Method\tablefootnote[6]{Results of Beyond-MSE, Conv-AE, and U-Net are from Liu \etal~\cite{liu2018ano_pred}.} & \data{Avenue} & \data{ShanghaiTech} & \data{Ped1} & \data{Ped2}\\
            \midrule
            Beyond-MSE~\cite{DBLP:journals/corr/MathieuCL15} & -- & --  & 0.200 & 0.396\\
            Conv-AE~\cite{Hasan_2016_CVPR} & 0.256 & -- & 0.243 & 0.384 \\
            U-Net~\cite{liu2018ano_pred} & 0.270 & -- & 0.243 & 0.435\\
            Liu \etal~\cite{liu2018ano_pred} & 0.275 & 0.175 & 0.259 & 0.469\\
            \cmidrule{1-5}
            \textbf{{\ourmethod}} (Ours) & \textbf{0.344} & \textbf{0.182} & \textbf{0.260} & \textbf{0.512}\\
            \bottomrule
        \end{tabular}
    }
\end{table}

\para{Case studies with visualizations}
In Fig.~\ref{fig:avenue-vis}, we show some examples of the anomaly score curves from our proposed method and plot some key frames with normal or anomalous events.
As shown in the figures clearly, our method responds to normal and anomalous events correctly.
The anomaly score increases drastically if an outlier suddenly intrudes and increases gradually if a person starts doing something unusual slowly.
If the objects causing the anomalies disappear in the frame, the anomaly score quickly decreases to a quite low level.

\begin{figure}[h]
    \centering
    \subfloat[Original Frame\label{fig:vis-module-a}]{
        \begin{minipage}[t]{0.23\columnwidth}
        \centering
        \includegraphics[width=0.9\linewidth]{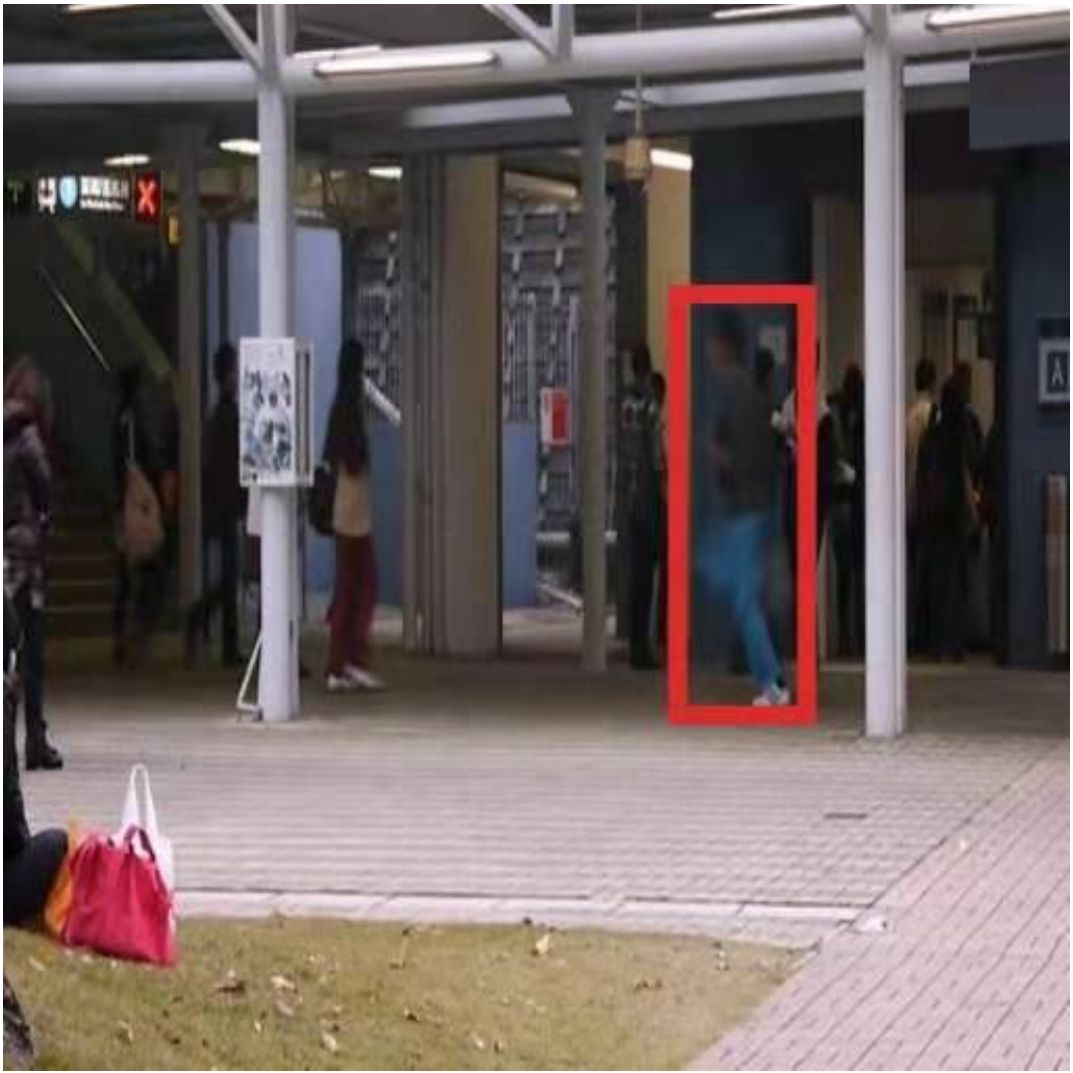}
        \end{minipage}
    }
    \subfloat[Encoder]{
        \begin{minipage}[t]{0.23\columnwidth}
        \centering
        \includegraphics[width=0.9\linewidth]{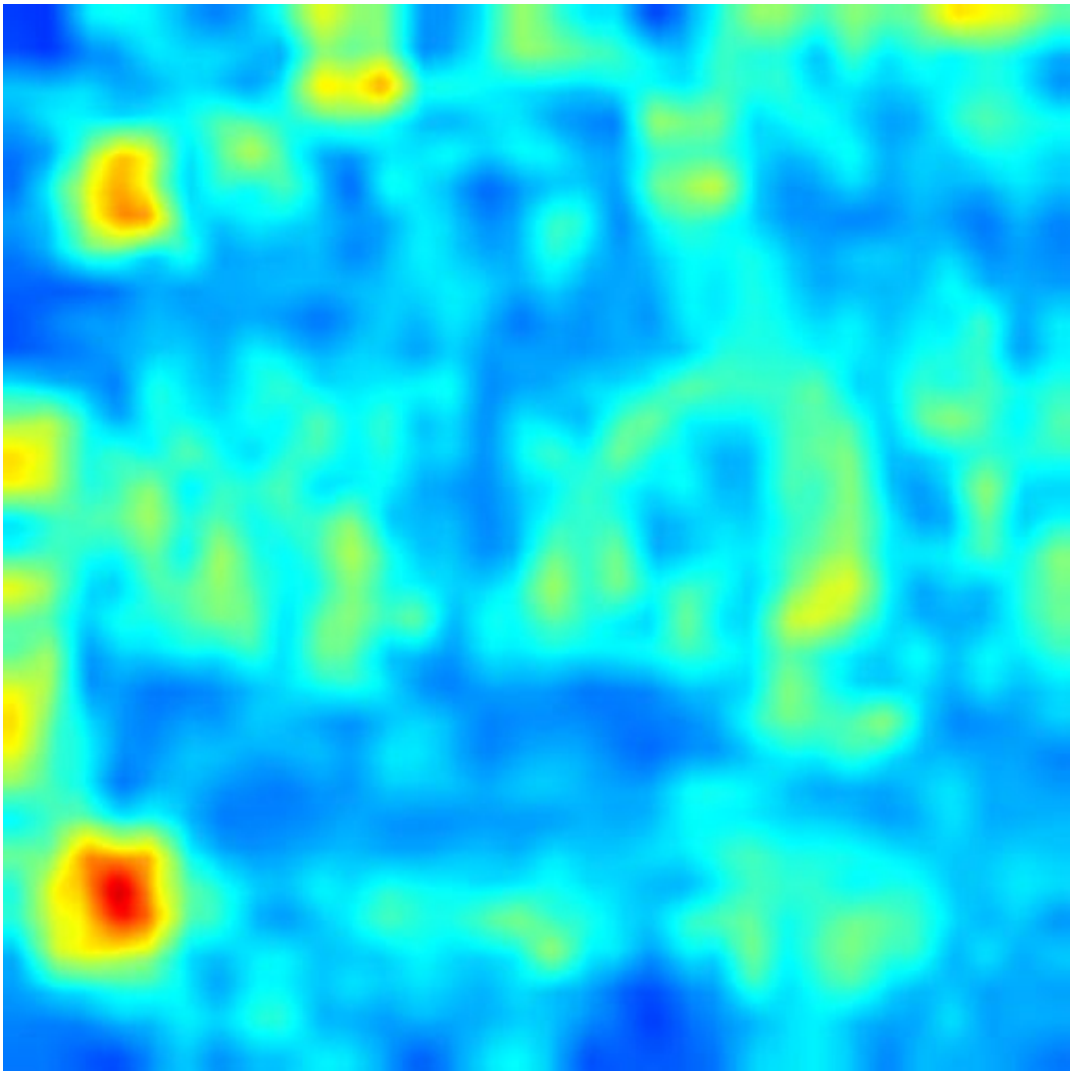}
        \end{minipage}
    }
    \subfloat[Non-local Block]{
        \begin{minipage}[t]{0.23\columnwidth}
        \centering
        \includegraphics[width=0.9\linewidth]{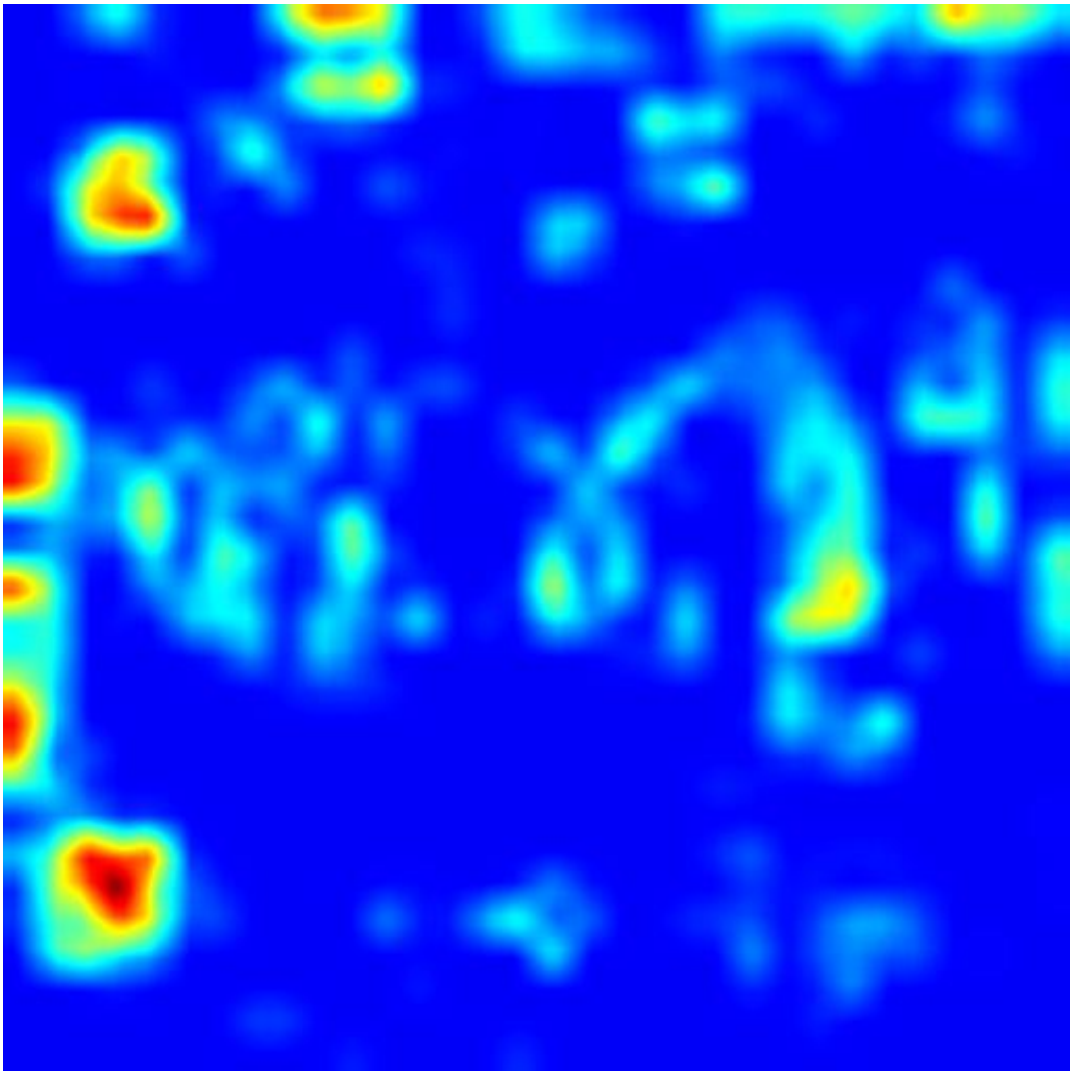}
        \end{minipage}
    }
    \subfloat[ConvGRU]{
        \begin{minipage}[t]{0.23\columnwidth}
        \centering
        \includegraphics[width=0.9\linewidth]{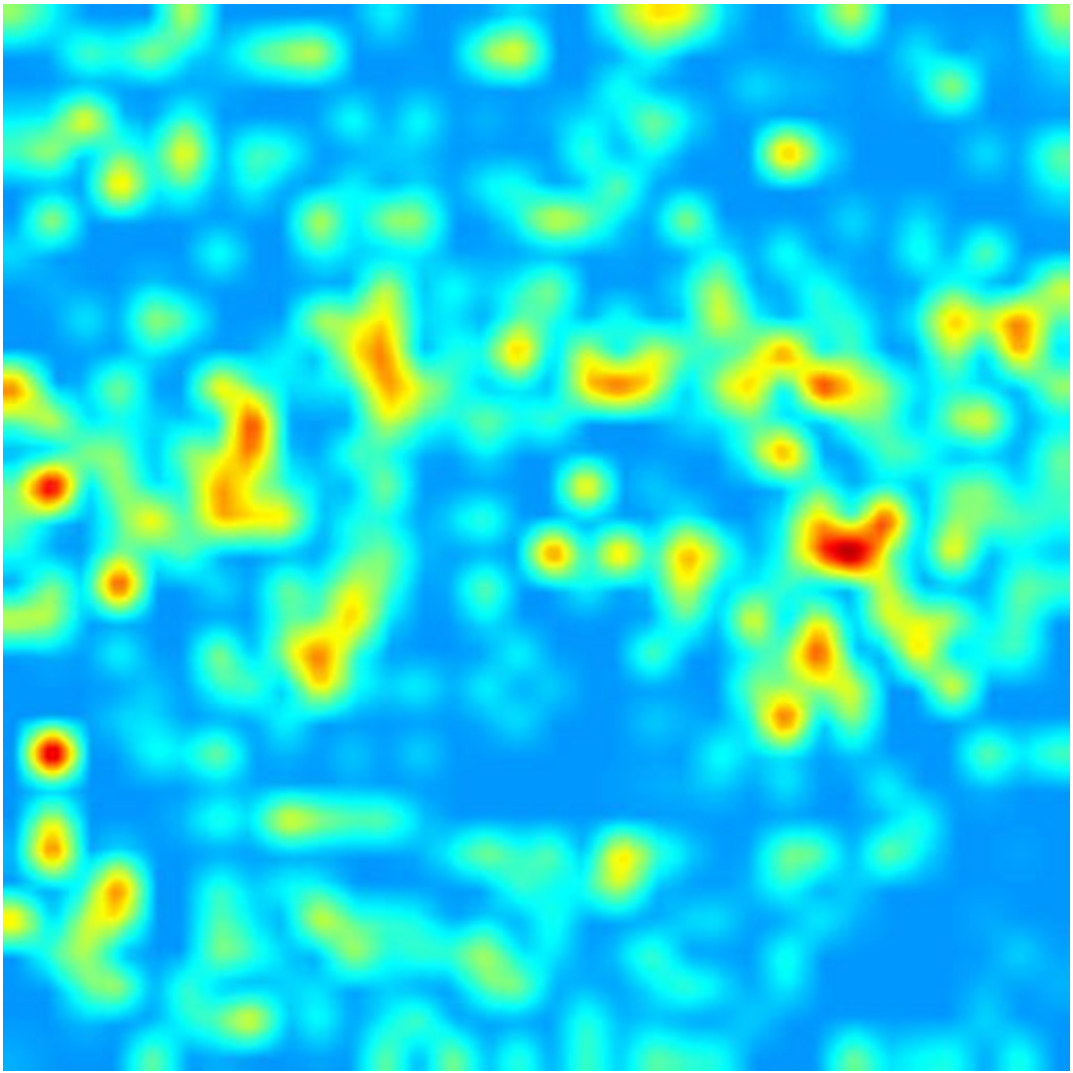}
        \end{minipage}
    }
    \caption{Visualizations of the original frame and the attentions of each component in path $h^{[3]}$.
    The abnormal frame shown in (a) is Frame 285 in \data{Avenue} Testing Video 02.}
    \label{fig:vis-module}
\end{figure}

\para{Investigations on the prediction module}
To investigate our proposed prediction module, we took an abnormal frame from \data{Avenue} and visually compared the components.
Fig.~\ref{fig:vis-module-a} shows the original video frame, where the anomaly event is ascribed to the person (in the red box) running from right to left.
Meanwhile, the people walking around and the bags on the grass are false clues to be excluded, as they appear in the entire video.
In the other three plots in Fig.~\ref{fig:vis-module}, the attentions of the encoder, the non-local block, and the ConvGRU in the deepest path $h^{[3]}$ are visualized by GRAD-CAM~\cite{selvaraju2017grad}.
The encoder itself aimed at transforming the original frame into the hidden states, and its attention was scattered over the image.
The non-local block, by extracting and exploiting long-range spatial contexts, had a condensed attention mainly focused on a few objects and locations of interests, such as the bags, the signboard, and the people.
The ConvGRU took the consecutive outputs from the non-local block and handled the temporal information. Therefore, it succeeded in excluding the static and normal objects while paying more attention to the people, especially the running person.
The attention visualizations clearly demonstrate the efficacy and the cooperation of the non-local and ConvGRU components in capturing the spatial and temporal dependencies and focusing on informative parts.

\begin{figure}[h]
    \centering
    \subfloat[Original Frame]{
        \begin{minipage}[t]{0.23\columnwidth}
        \centering
        \includegraphics[width=0.9\linewidth]{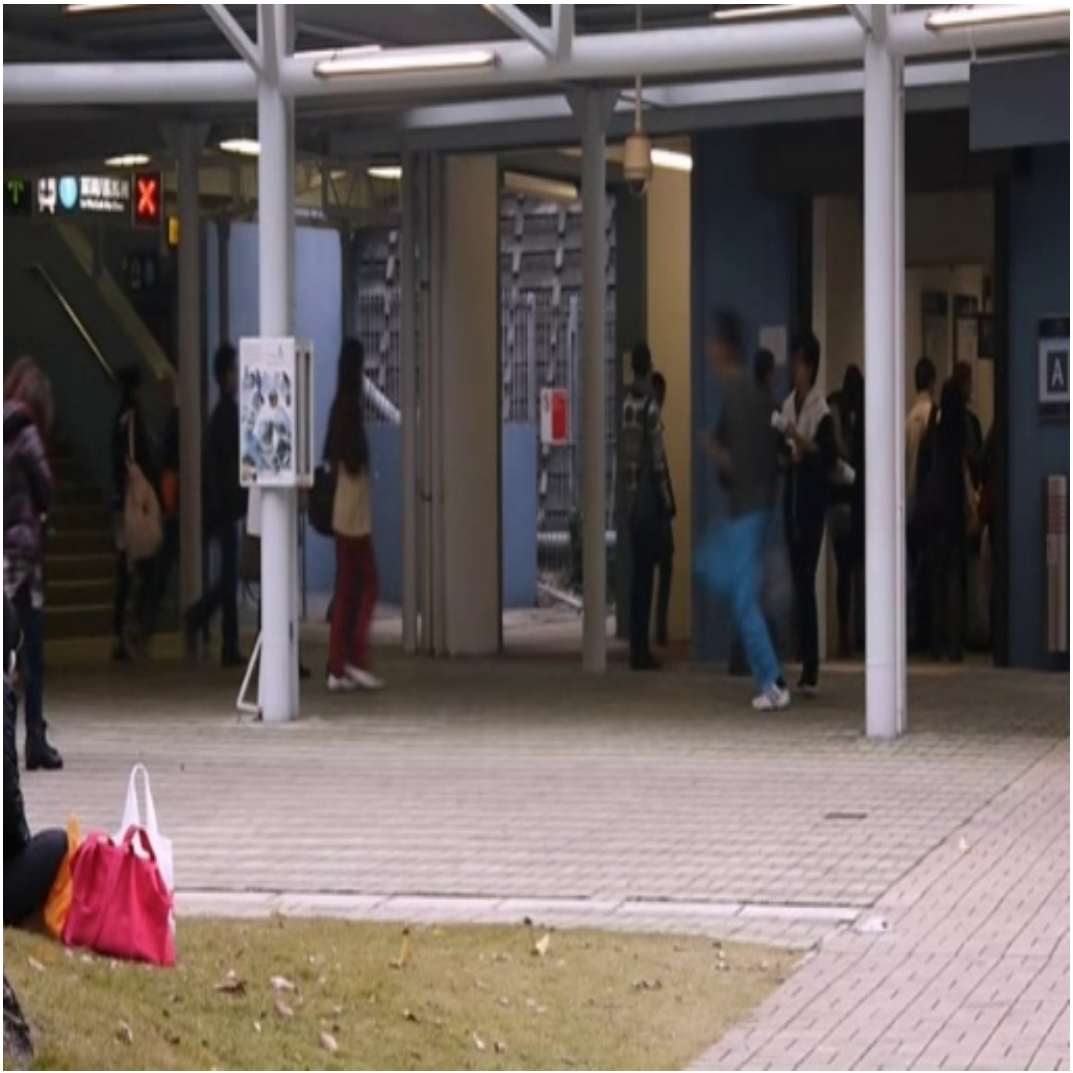}
        \end{minipage}
    }
    \subfloat[Path $h^{[1]}$]{
        \begin{minipage}[t]{0.23\columnwidth}
        \centering
        \includegraphics[width=0.9\linewidth]{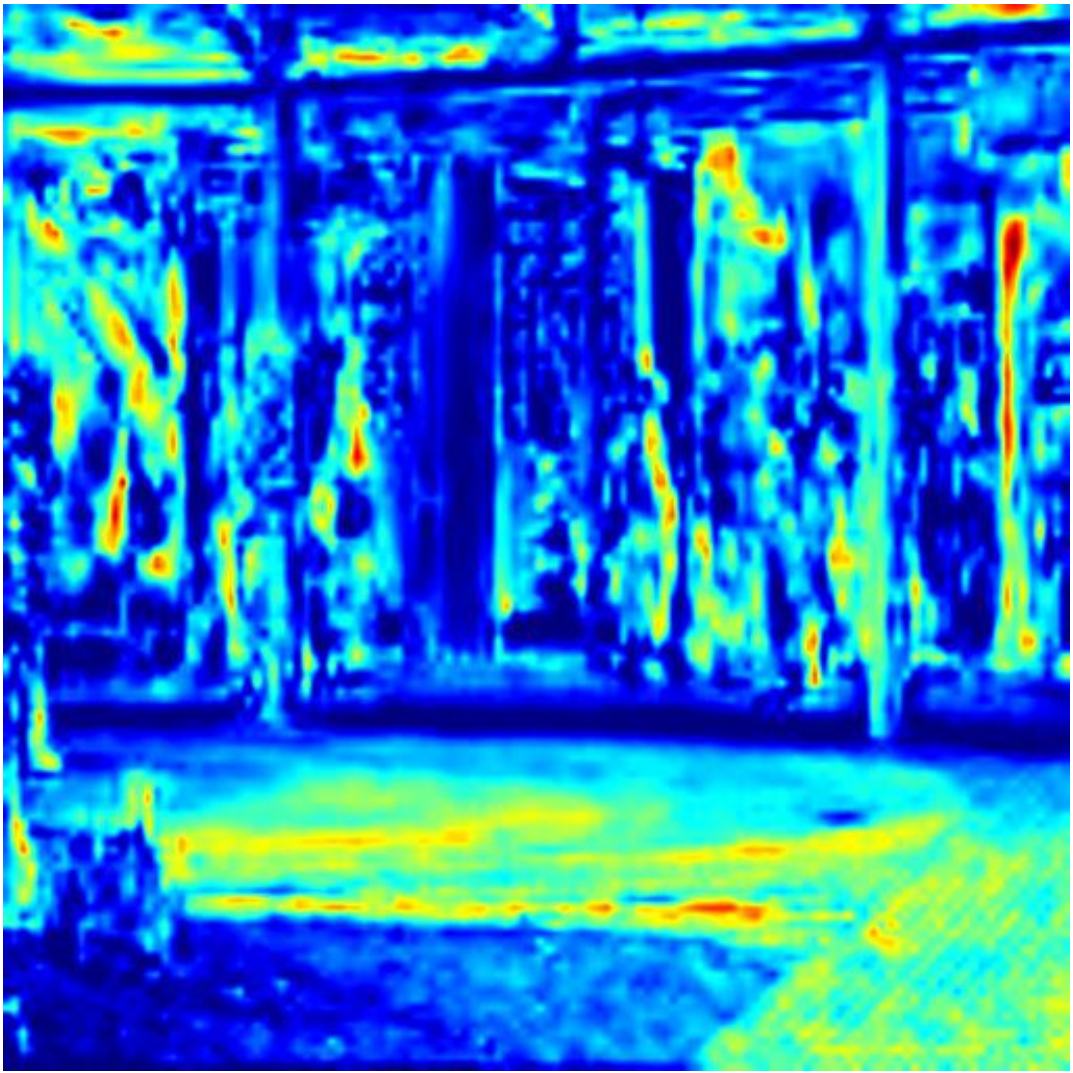}
        \end{minipage}
    }
    \subfloat[Path $h^{[2]}$]{
        \begin{minipage}[t]{0.23\columnwidth}
        \centering
        \includegraphics[width=0.9\linewidth]{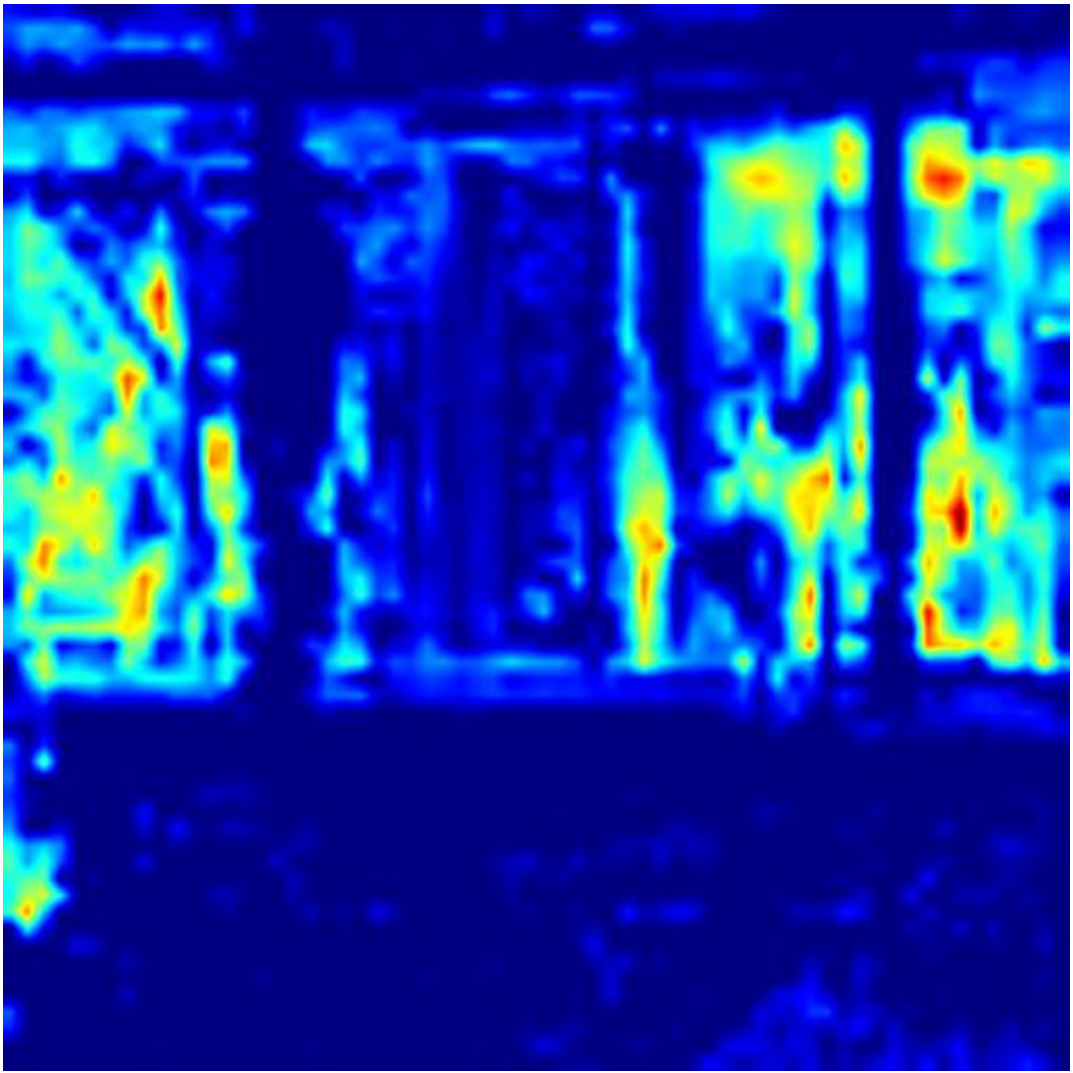}
        \end{minipage}
    }
    \subfloat[Path $h^{[3]}$]{
        \begin{minipage}[t]{0.23\columnwidth}
        \centering
        \includegraphics[width=0.9\linewidth]{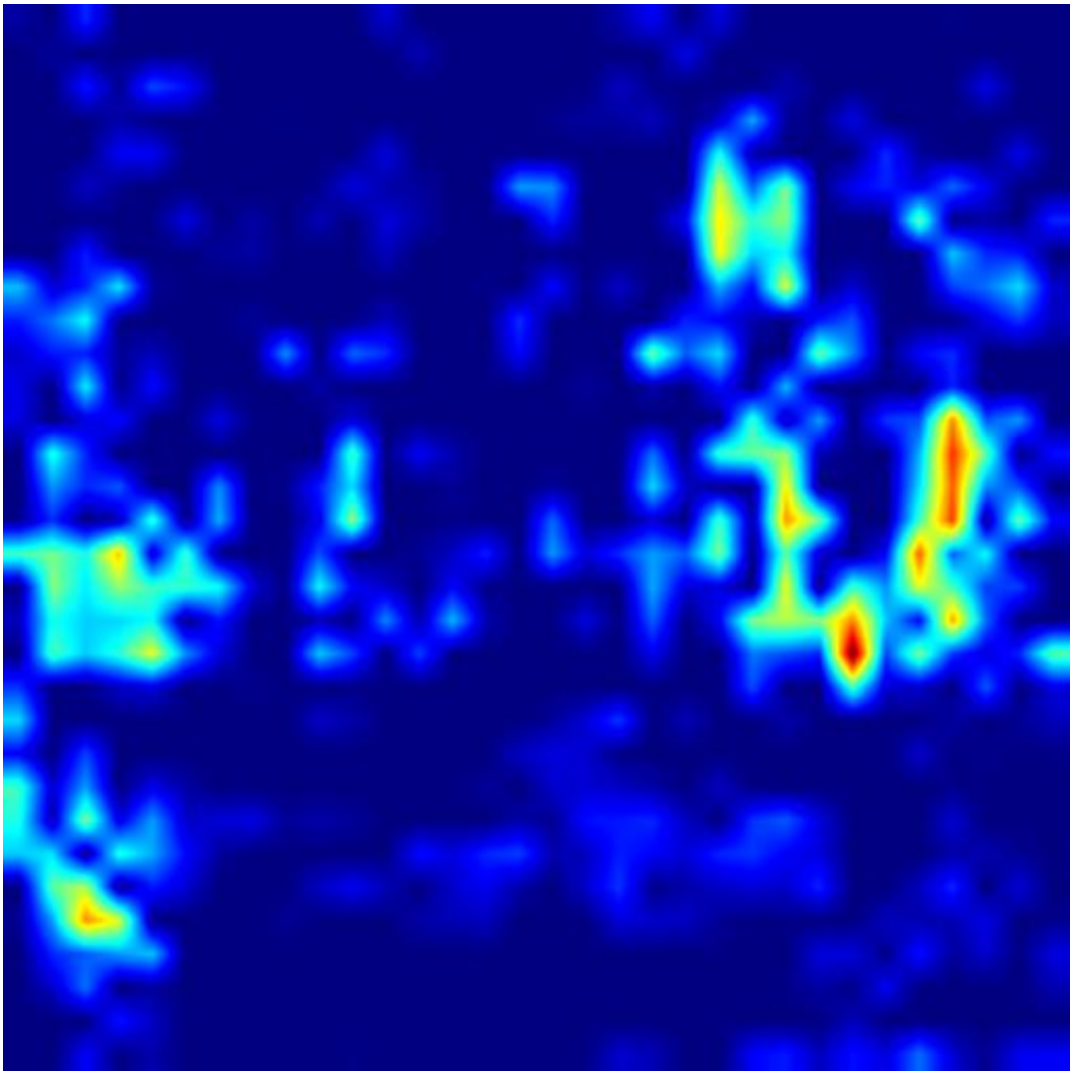}
        \end{minipage}
    }
    \caption{Visualizations of the original frame and the outputs from multi-path structure before the decoder.
    The abnormal frame shown in (a) is Frame 285 in \data{Avenue} Testing Video 02.}
    \label{fig:vis-multipath}
\end{figure}

\para{Investigations on the multi-path structure}
We took the same abnormal frame to further investigate the behaviors of the multi-path structure.
Fig.~\ref{fig:vis-multipath} shows the output from each prediction path before the decoder (by averaging the values from all channels).
As we expected, the model mainly used the shallow path $h^{[1]}$ to encode the invariant frame background such as the ground and road.
The middle path $h^{[2]}$ focused on more subtle parts of the frame such as the people;
And the deep path $h^{[3]}$ paid extra attention to the challenging parts, e.g., the blurred areas and the bags of different colors.
This showcase demonstrates the multi-path architecture, by capturing multi-scale spatial information, can pay more attention to semantically informative parts and less to background parts.

\para{Evaluations by AUPRC and F1 scores}
In real-world scenarios, anomaly events are expected to occur rarely, and the anomaly detection is usually a problem with imbalanced data.
In these settings, the metrics such as AUPRC or F1 scores are more appropriate and reliable.
Among the four benchmark datasets, only \data{Avenue} and \data{ShanghaiTech} have fewer abnormal frames than normal ones.
We calculated the AUPRC score and F1 scores (Max-F1 and F1$_{0.5}$) on these two datasets.
For the Max-F1 score, we varied the threshold on the anomaly score and reported the maximum F1 score. The best thresholds of our method are 0.541 and 0.377 for \data{Avenue} and \data{ShanghaiTech}, respectively.
For the F1$_{0.5}$ score, we simply set the threshold to be 0.5 and calculated the F1 score.
As shown in Table~\ref{tab:prc}, these new metrics further validate the superiority of our proposed method.
Meanwhile, the values of F1$_{0.5}$ are close to the max-F1 scores (only 0.1 and 0.4 percentage point difference on \data{Avenue} and \data{ShanghaiTech}, respectively), which indicates that the performance of our method is not too sensitive to the threshold values, and a threshold of 0.5 is moderate.

\begin{table}[h]
    \centering
    \caption{
        AUPRC and F1 scores of the anomaly detection results.
    }
    \label{tab:prc}
    {
        \begin{tabular}{l@{\ \ }c@{\ }c@{\ }c@{\ \ }c@{\ }c@{\ }c}
            \toprule
             &  \multicolumn{3}{c}{\data{Avenue}} &  \multicolumn{3}{c}{\data{ShanghaiTech}} \\
            \cmidrule(lr){1-1}  \cmidrule(lr){2-4} \cmidrule(lr){5-7}
            \multicolumn{1}{c}{Score} & AUPRC & Max-F1 & F1$_{0.5}$  & AUPRC & Max-F1 & F1$_{0.5}$ \\
            \midrule
            Liu \etal~\cite{liu2018ano_pred} & 93.4\% & 88.1\% & 88.0\% & -- & -- & -- \\
            \cmidrule{1-7}
            \textbf{{\ourmethod}} (Ours) & \textbf{94.4\%} & \textbf{89.4\%} & \textbf{89.3\%} & \textbf{72.4\%} & \textbf{74.5\%} & \textbf{74.1\%}\\
            \bottomrule
        \end{tabular}
    }
\end{table}

\subsection{Discussions about Performance Trade-off}
With its default configuration, our method can obtain superb anomaly detection performance with a high speed of 0.109 seconds per frame, with its moderate model size of 102MB and executing memory of 619MB.
We proposed and analyzed a few strategies on hyperparameter selection and model inference with the main purpose on model efficiency and detection accuracy trade-off.

\para{Comparisons on numbers of input frames}
Hyperparameters such as the number of input frames $P$ can be easily adjusted to trade off between running speed and detection quality.
Recent work took different numbers of input frames (\eg, 16 frames~\cite{Abati_2019_CVPR,Gong_2019_ICCV} or 4 frames~\cite{liu2018ano_pred}),
and we trained and evaluated our method with different values of $P$ and reported their AUROC score differences from the initial setting $P=8$ in Table~\ref{tab:length} as well as the overall anomaly detection time.
Using more input frames ($>8$) only gained limited improvement on \data{Avenue} of no more than 0.1 percentage points in terms of the AUROC score and even worse performance on \data{ShanghaiTech}.
The results indicated that $P=8$ is a proper choice with sufficiently fast running speed of about 100 milliseconds per frame.

\begin{table}[h]
    \centering
    \caption{
        Performance comparisons on \data{Avenue} and \data{ShanghaiTech} with different input lengths $P$. AUROC score increase/decrease compared with $P=8$ in terms of percentage points is shown.
    }
    \label{tab:length}
    {
        \begin{tabular}{cccc}
            \toprule
            \multirow{2.5}{*}{\shortstack{\# of Input Frames\\$P$}}
             & \multicolumn{2}{c}{AUROC Score Gain} &
             \multirow{2.5}{*}{\shortstack{Average Running\\Time (s)}} \\
            \cmidrule(lr){2-3}
             & \data{Avenue} & \data{ShanghaiTech} & \\
            \midrule
            4 & $-$3.4\% & $-$2.4\% & 0.065 \\
            6 & $-$0.6\% & $-$1.3\% & 0.086 \\
            8 & \textit{(88.3\%)} & \textit{(76.6\%)} & 0.109 \\
            10 & $+$0.1\%  &  $-$0.2\% & 0.132 \\
            12 & $+$0.1\%  & $-$1.0\% & 0.156 \\
            16 & $+$0.0\%  & $-$2.5\% & 0.201 \\
            \bottomrule
        \end{tabular}
    }
\end{table}

\para{Fast detection by reusing previous outputs}
Another idea for fast anomaly detection is to reuse the predictor output $ \vct{h}_t$ calculated at time $t$ in later time instead of recalculating it.
Recall that, in our method, $P$ input frames were used for anomaly detection of the next one frame.
We could feed $P+Q-1$ frames into our model and conduct anomaly detection for $Q$ frames with $Q>1$, even though the model was trained with inputs of fixed-length $P=8$.
As shown in Table~\ref{tab:slice}, larger $Q$ saved more time while suffering the problem of error accumulation, possibly from longer RNNs.
However, even with the standard real-time requirement ($>$30 fps) in video surveillance, our method still maintained an AUROC score of 87.5\% (with $Q=24$ and the speed of 32.2 fps) on \data{Avenue} which is better than all other baseline results in Table~\ref{tab:scores}. On \data{ShanghaiTech}, the performance degradation is slightly larger but the results with $Q\le8$ are still competitive.

\begin{table}[h]
    \centering
    \caption{
        Performance comparisons with different prediction lengths $Q$ during testing. AUROC score increase/decrease compared with $Q=1$ in terms of percentage points is shown.
    }
    \label{tab:slice}
    {
        \begin{tabular}{cccc}
            \toprule
            \multirow{2.5}{*}{\shortstack{\# of Input\\to Predict $Q$}}
             & \multicolumn{2}{c}{AUROC Score Gain} &
             \multirow{2.5}{*}{\shortstack{Amortized Running\\Time (s)}} \\
            \cmidrule(lr){2-3}
             & \data{Avenue} & \data{ShanghaiTech} & \\
            \midrule
            1 & \textit{(88.3\%)} & \textit{(76.6\%)} & 0.109 \\
            4 & $+$0.1\% & $-$0.1\% & 0.064 \\
            8 & $-$0.1\% & $-$0.8\% & 0.045 \\
            16 & $-$0.7\% & $-$1.7\% & 0.036 \\
            24 & $-$0.8\% & $-$1.9\% & 0.031 \\
            32 & $-$1.9\% & $-$2.3\% & 0.030 \\
            48 & $-$1.7\% & $-$3.4\% & 0.028 \\
            \bottomrule
        \end{tabular}
    }
\end{table}

\para{Comparisons with different input sizes}
The input size is another factor affecting the model efficiency.
In the main experiments, we chose the input size 256$\times$256 as other baselines~\cite{liu2018ano_pred,ye2019anopcn} did for a fair comparison. Additionally, we evaluated different input sizes on the proposed model, which was trained on inputs of 256$\times$256.
As shown in Table~\ref{tab:inputsize}, small input size decreased the detection performance drastically, e.g., $>5$ percentage points drop with the halved height and width. As the proposed model trained with the input size 256$\times$256 only has three ConvGRU paths with 8$\times$ downsampling, it performed slightly worse on larger input size with a doubled running time.

\begin{table}[h]
    \centering
    \caption{
        Performance comparisons with different input frame sizes.
        AUROC score increase/decrease compared with input size 256 $\times$ 256 in terms of percentage points is shown.
    }
    \label{tab:inputsize}
    {
        \begin{tabular}{cccc}
            \toprule
            \multirow{2.5}{*}{Input Frame Size}
             & \multicolumn{2}{c}{AUROC Score Gain} &
             \multirow{2.5}{*}{\shortstack{Average Running\\Time (s)}} \\
            \cmidrule(lr){2-3}
             & \data{Avenue} & \data{ShanghaiTech} & \\
            \midrule
            128 $\times$ 128 & $-$6.0\% & $-$5.4\% & 0.049 \\
            192 $\times$ 192 & $-$3.1\% & $-$2.4\% & 0.072 \\
            256 $\times$ 256 & \textit{(88.3\%)} & \textit{(76.6\%)} & 0.109 \\
            384 $\times$ 384 & $-$1.0\% & $-$2.1\% & 0.216 \\
            \bottomrule
        \end{tabular}
    }
\end{table}

\section{Conclusion}
In this paper, we proposed {\ourmethod}, a novel unsupervised video anomaly detection method based on frame prediction.
We designed its {\multi} predictor which is more suitable for distinguishing normal and abnormal events in videos.
we introduced a {\ntloss} in model training to mitigate the interference of noise to the prediction network and improve the robustness of the anomaly detection.
The experimental results on three datasets as well as exhaustive ablation studies and investigations validated the effectiveness of our design and the superiority of the proposed method.

\ifCLASSOPTIONcaptionsoff
  \newpage
\fi

\bibliographystyle{IEEEtran}
\bibliography{Robust_Unsupervised_Video_Anomaly_Detection}

\newpage

\begin{IEEEbiography}[{\includegraphics[width=1in,height=1.25in,clip,keepaspectratio]{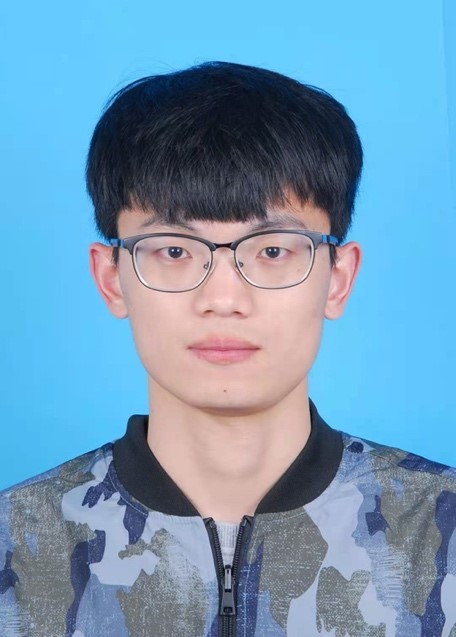}}]{Xuanzhao Wang}
Xuanzhao Wang received the B.S. degree from the Beijing University of Posts and Telecommunications in 2017, where he is currently pursuing the M.S. degree. His research interests include object tracking, anomaly detection, and deep learning.
\end{IEEEbiography}

\vspace{-0.3in}

\begin{IEEEbiography}[{\includegraphics[width=1in,height=1.25in,clip,keepaspectratio]{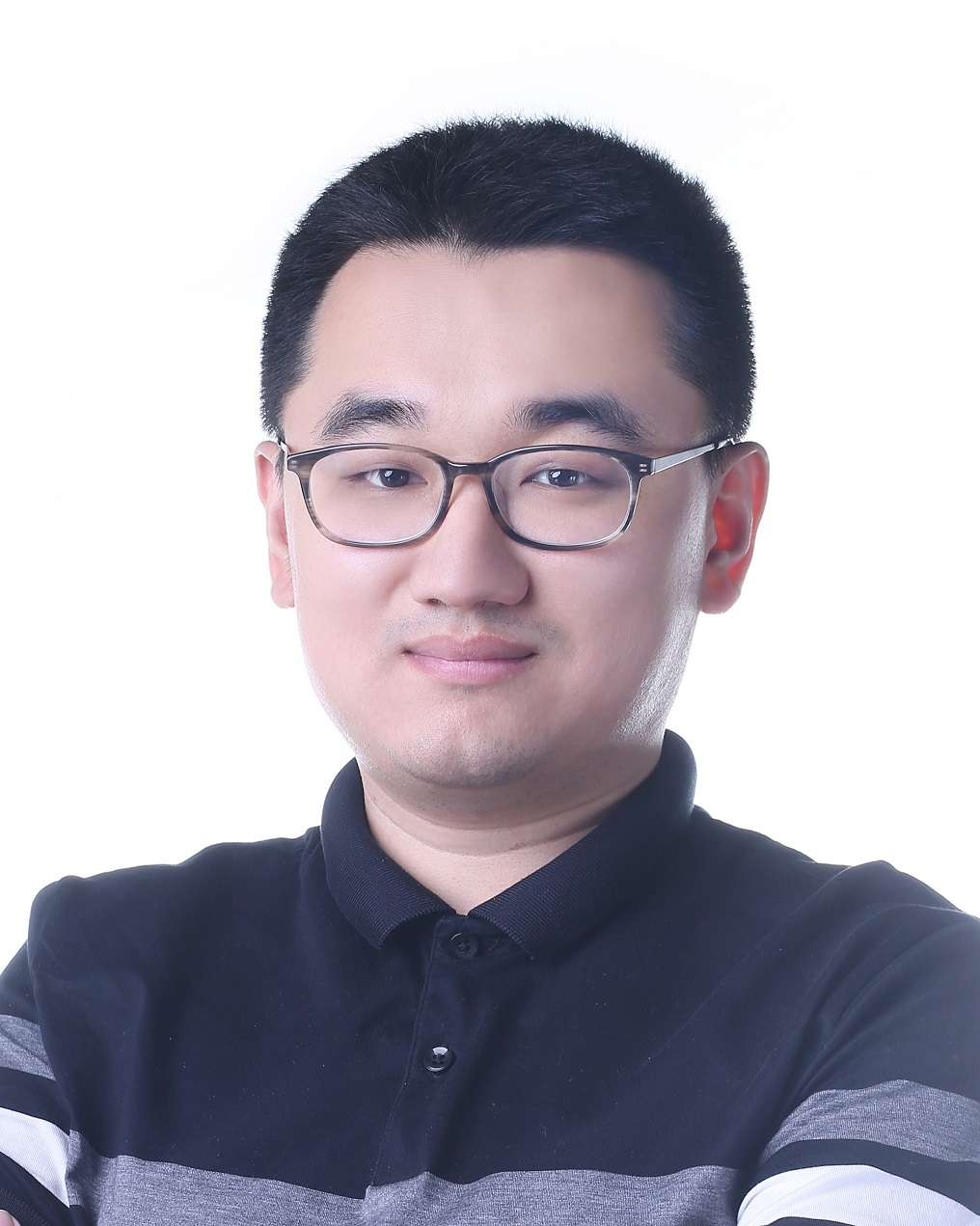}}]{Zhengping Che}
Zhengping Che received the Ph.D. degree in Computer Science from the University of Southern California, Los Angeles, CA, USA, in 2018, and the B.Eng. degree in Computer Science from Tsinghua University, Beijing, China, in 2013.
He is a research scientist at Didi Chuxing.
His current research interests lie in machine learning, deep learning, and artificial intelligence.
\end{IEEEbiography}

\vspace{-0.3in}

\begin{IEEEbiography}[{\includegraphics[width=1in,height=1.25in,clip,keepaspectratio]{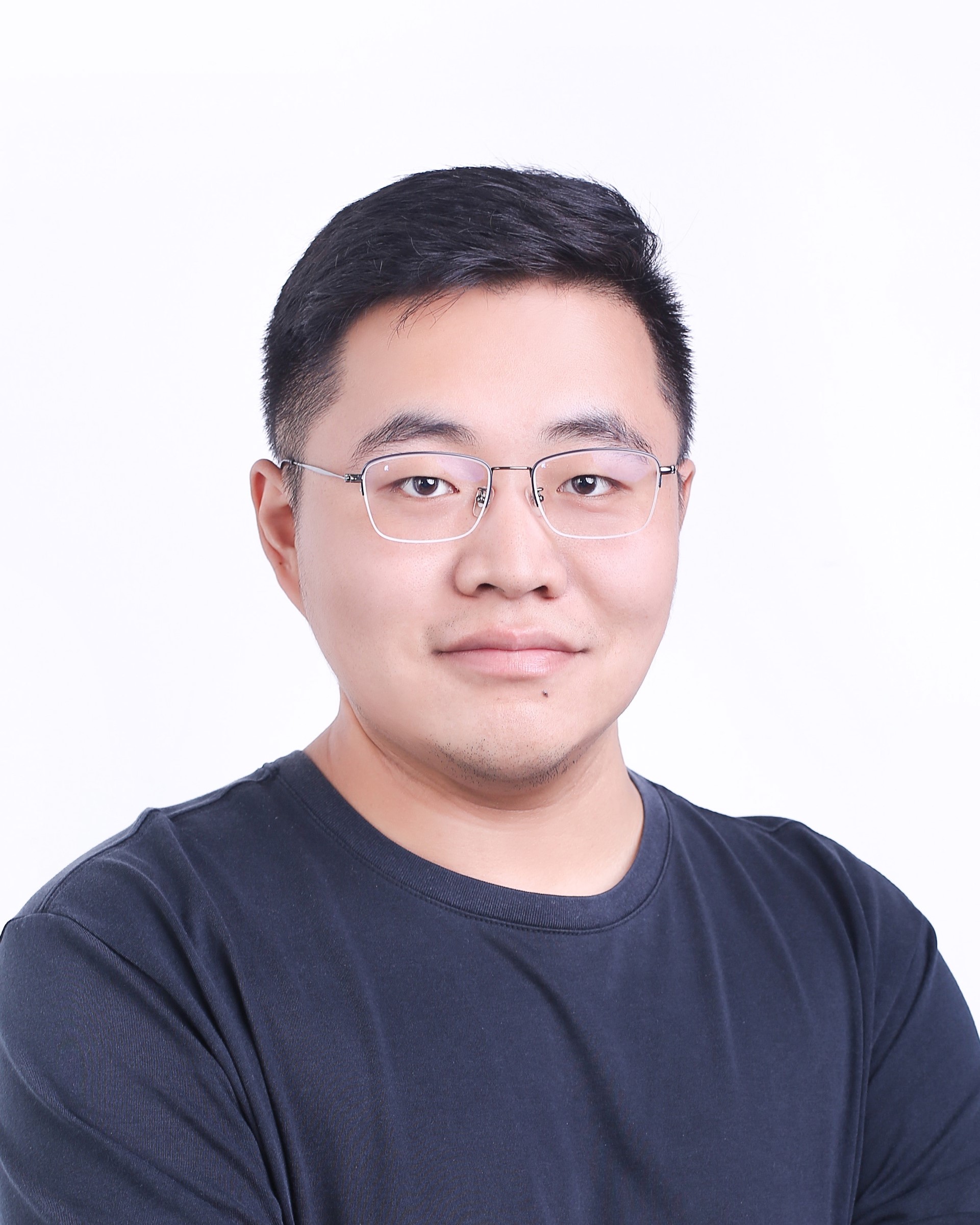}}]{Bo Jiang}
Bo Jiang received the master degree from the Department of Computer Science, University of Southern California.
He is a research scientist at Didi Chuxing. His research interests include machine learning and deep learning.
\end{IEEEbiography}

\vspace{-0.3in}

\begin{IEEEbiography}[{\includegraphics[width=1in,height=1.25in,clip,keepaspectratio]{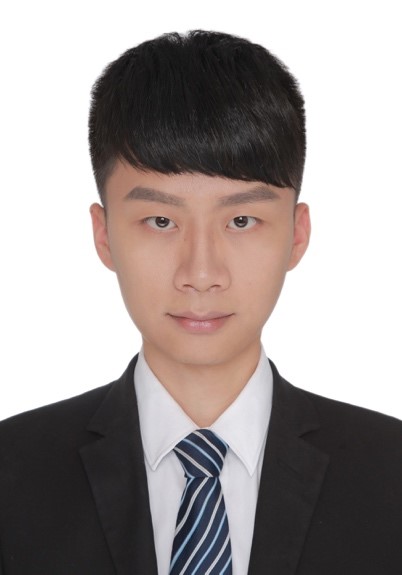}}]{Ning Xiao}
Ning Xiao received the B.E. degree from the Beijing University of Posts and Telecommunications in 2018, where he is currently pursuing the M.E. degree.
His current research interests include deep learning, object detection, computer vision, and image processing.
\end{IEEEbiography}

\vspace{-0.3in}

\begin{IEEEbiography}[{\includegraphics[width=1in,height=1.25in,clip,keepaspectratio]{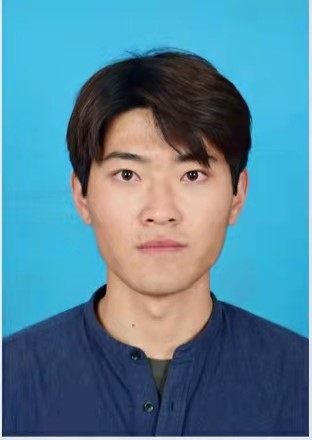}}]{Ke Yang}
Ke Yang received the B.S. degree from the Beijing University of Posts and Telecommunications in 2017, where he is currently pursuing the M.S. degree. His research interests include artificial intelligence, neural networks, and action recognition.
\end{IEEEbiography}

\newpage

\begin{IEEEbiography}[{\includegraphics[width=1in,height=1.25in,clip,keepaspectratio]{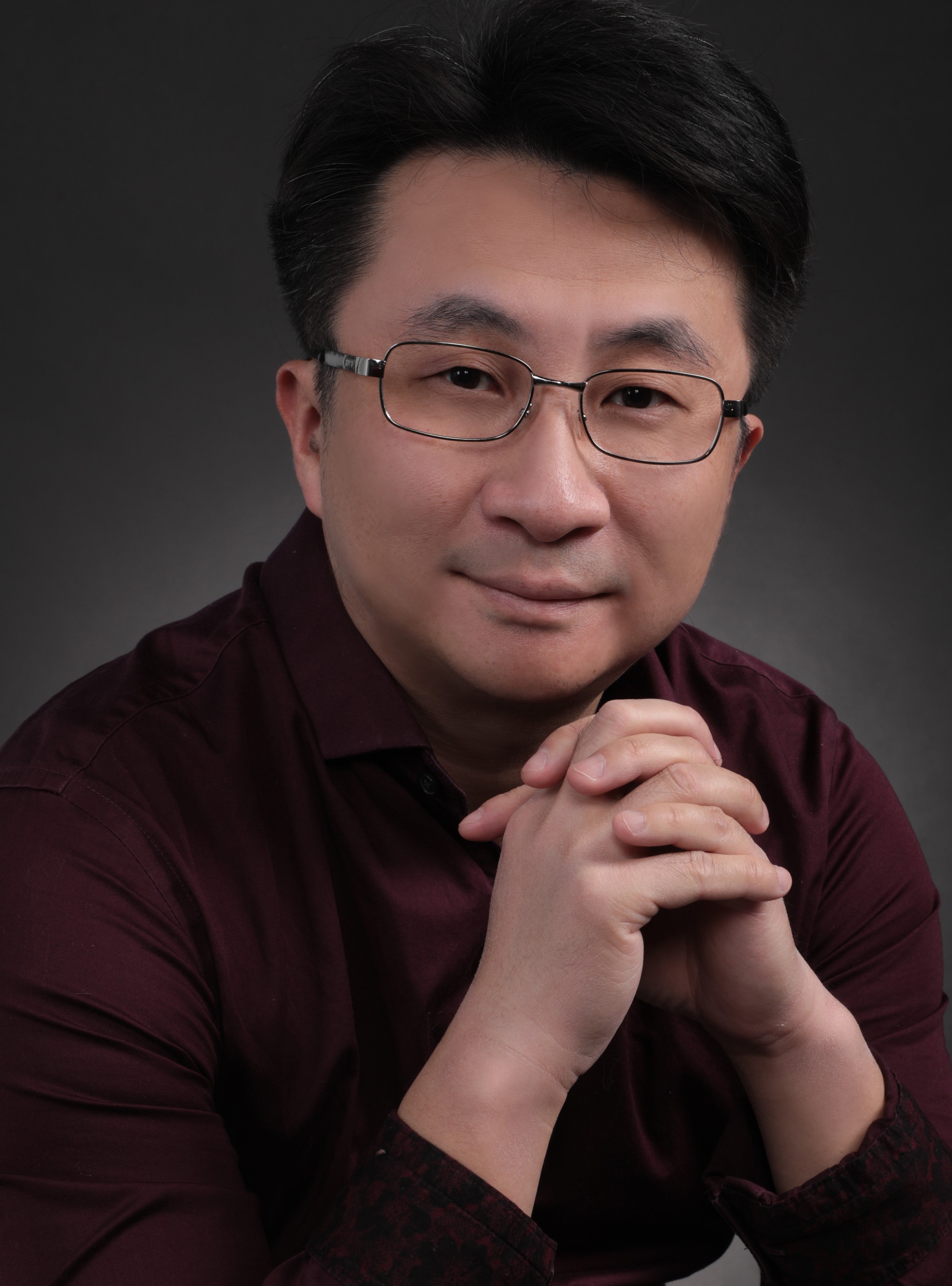}}]{Jian Tang}
Dr. Jian Tang received his Ph.D degree in Computer Science from Arizona State University in 2006. He is an IEEE Fellow and an ACM Distinguished Member. He is with Midea Group. His research interests lie in the areas of AI, IoT, Wireless Networking, Mobile Computing and Big Data Systems. Dr. Tang has published over 160 papers in premier journals and conferences. He received an NSF CAREER award in 2009. He also received several best paper awards, including the 2019 William R. Bennett Prize and the 2019 TCBD (Technical Committee on Big Data) Best Journal Paper Award from IEEE Communications Society (ComSoc), the 2016 Best Vehicular Electronics Paper Award from IEEE Vehicular Technology Society (VTS), and Best Paper Awards from the 2014 IEEE International Conference on Communications (ICC) and the 2015 IEEE Global Communications Conference (Globecom) respectively. He has served as an editor for several IEEE journals, including IEEE Transactions on Big Data, IEEE Transactions on Mobile Computing, etc. In addition, he served as a TPC co-chair for a few international conferences, including the IEEE/ACM IWQoS’2019, MobiQuitous’2018, IEEE iThings’2015. etc.; as the TPC vice chair for the INFOCOM’2019; and as an area TPC chair for INFOCOM 2017-2018. He is also an IEEE VTS Distinguished Lecturer, and the Chair of the Communications Switching and Routing Committee of IEEE ComSoc.
\end{IEEEbiography}

\vspace{-0.3in}

\begin{IEEEbiography}[{\includegraphics[width=1in,height=1.25in,clip,keepaspectratio]{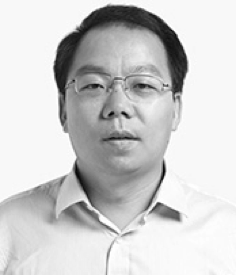}}]{Jieping Ye}
Jieping Ye received the Ph.D. degree in computer science from the University of Minnesota, Twin Cities, MN, USA, in 2005. He is currently with the Beike AI Tech and also an Associate Professor with the University of Michigan, Ann Arbor, MI, USA. Before that he was the VP of Didi Chuxing and a DiDi Fellow. His research interests include big data, machine learning, and data mining, with applications in transportation and biomedicine. He was a recipient of the NSF CAREER Award in 2010. His papers have been selected for the Outstanding Student Paper at ICML in 2004, the KDD Best Research Paper Runner Up in 2013, and the KDD Best Student Paper Award in 2014. He has served as a Senior Program Committee/Area Chair/Program Committee Vice Chair for many conferences, including NeurIPS, ICML, KDD, IJCAI, ICDM, and SDM. He has served as an Associate Editor for \textit{Data Mining and Knowledge Discovery}, the \textit{IEEE Transactions on Knowledge and Data Engineering}, and the \textit{IEEE Transactions on Pattern Analysis and Machine Intelligence}. He is a Fellow member of the IEEE.
\end{IEEEbiography}

\vspace{-0.3in}

\begin{IEEEbiography}[{\includegraphics[width=1in,height=1.25in,clip,keepaspectratio]{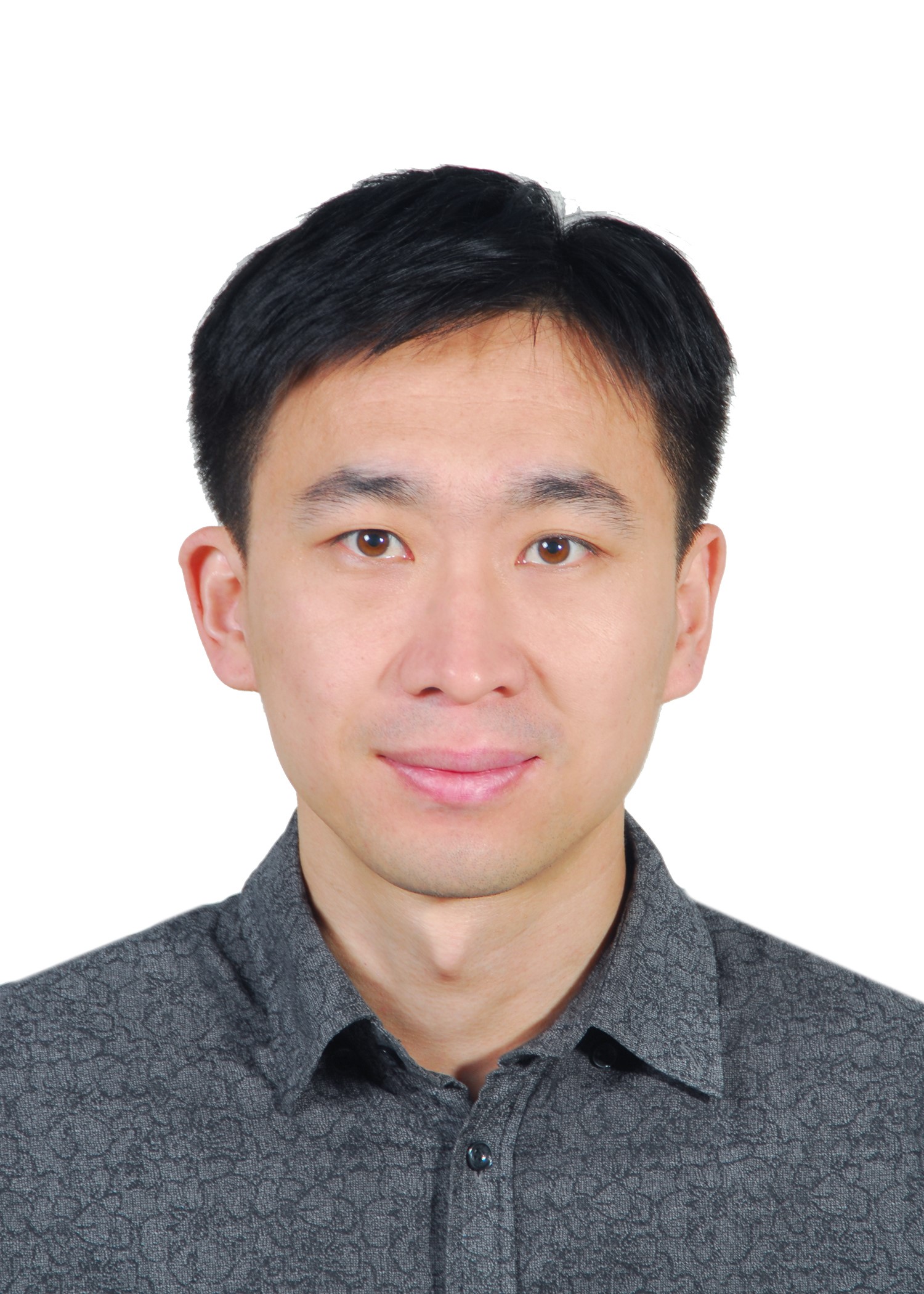}}]{Jingyu Wang}
Jingyu Wang obtained his PhD degree from Beijing University of Posts and Telecommunications in 2008.
He is currently a professor of State Key Laboratory of Networking and Switching Technology at Beijing University of Posts and Telecommunications.
He has publishes more than 50 papers in international journal, including IEEE COMMAG, TCC, TWC, TMM, TVT, and so on.
His research interests span broad aspects of SDN, big data processing and transmission technology, overlay networks, multi-media services and communication, and traffic engineering.
\end{IEEEbiography}

\vspace{-0.3in}

\begin{IEEEbiography}[{\includegraphics[width=1in,height=1.25in,clip,keepaspectratio]{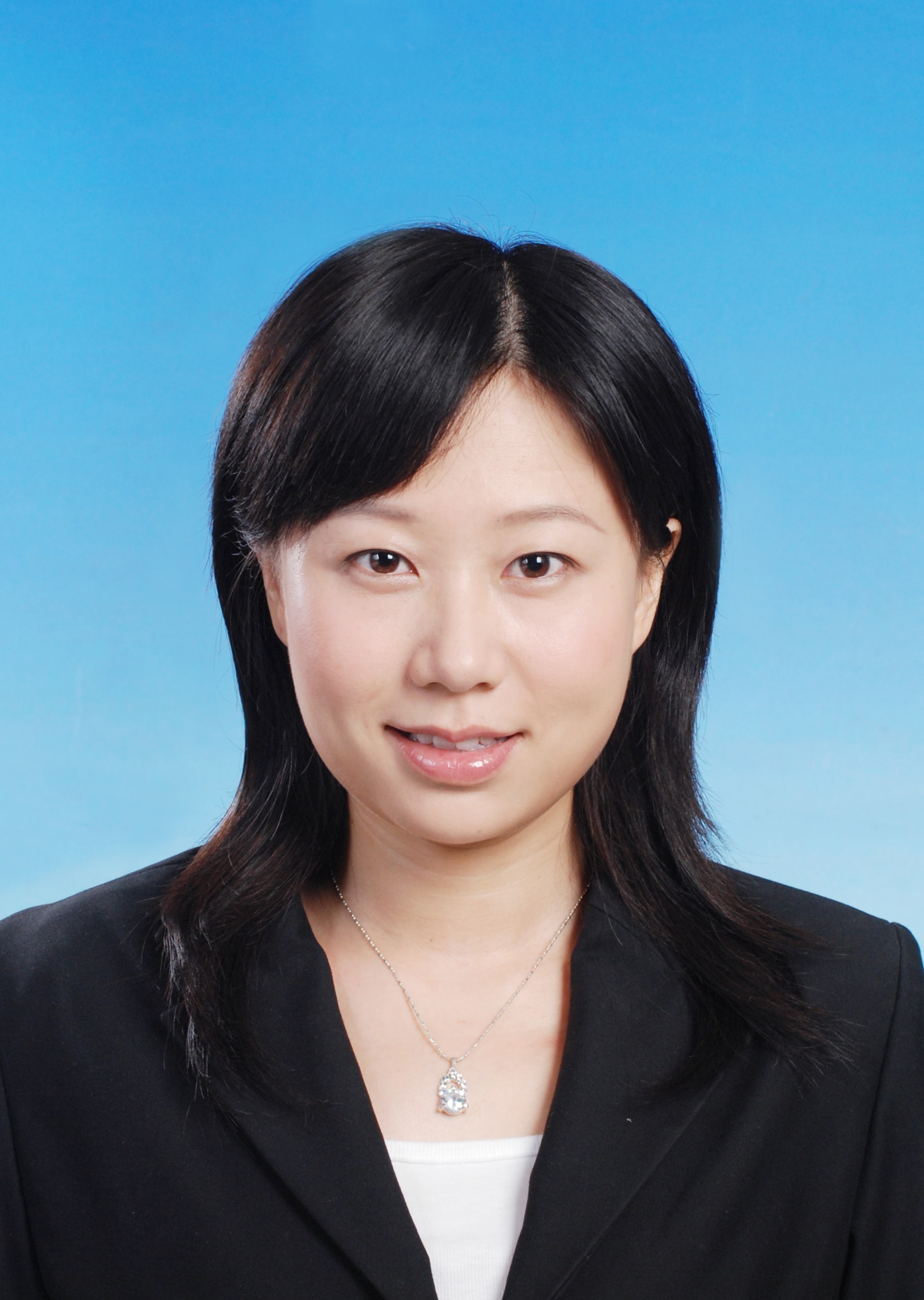}}]{Qi Qi}
Qi Qi obtained her PhD degree from Beijing University of Posts and Telecommunications in 2010.
She is an associate professor of State Key Laboratory of Networking and Switching Technology at Beijing University of Posts and Telecommunications.
She has published more than 30 papers in international journal, and obtained two National Natural Science Foundations of China.
Her research interests include edge computing, mobile cloud computing, Internet of Things, ubiquitous services, deep learning, and deep reinforcement learning.
\end{IEEEbiography}

\end{document}